%% file: neurips_2025.tex
\documentclass{article}
\PassOptionsToPackage{numbers}{natbib}

\usepackage[preprint]{neurips_2025}

\usepackage[utf8]{inputenc} 
\usepackage[T1]{fontenc}    
\usepackage{hyperref}       
\usepackage{url}            
\usepackage{booktabs}       
\usepackage{amsfonts}       
\usepackage{nicefrac}       
\usepackage{microtype}      
\usepackage{xcolor}         

\usepackage{graphicx}
\usepackage{booktabs}
\usepackage{multirow}
\usepackage{pifont}
\usepackage{subcaption}
\usepackage{amsmath}
\usepackage{wrapfig}
\usepackage{multirow}
\usepackage{makecell}
\usepackage[utf8]{inputenc}
\usepackage[most]{tcolorbox}
\usepackage{amsfonts}
\usepackage{enumitem}

\title{Revisiting Uncertainty Estimation and Calibration of Large Language Models}
\author{%
  Linwei Tao\\
  School of Computer Science\\
  University of Sydney\\
  \texttt{linwei.tao@sydney.edu.au} \\
  \And
  Yi-Fan Yeh \\
  School of Computer Science\\
  University of Sydney\\
  \texttt{yyeh7345@uni.sydney.edu.au} \\
  \And
  Minjing Dong \\
  City University of Hong Kong \\
  \texttt{minjdong@cityu.edu.hk} \\
  \And
  Tao Huang \\
  Shanghai Jiao Tong University \\
  \texttt{t.huang@sjtu.edu.cn} \\
  \And
  Philip Torr \\
  Department of Engineering Science\\
  University of Oxford \\
  \texttt{philip.torr@eng.ox.ac.uk} \\
  \And
  Chang Xu \\
  School of Computer Science\\
  University of Sydney\\
  \texttt{c.xu@sydney.edu.au} \\
}

\begin{document}

\maketitle

\begin{abstract}
As large language models (LLMs) are increasingly deployed in high-stakes applications, robust uncertainty estimation is essential for ensuring the safe and trustworthy deployment of LLMs. We present the most comprehensive study to date of uncertainty estimation in LLMs, evaluating 80 models spanning open- and closed-source families, dense and Mixture-of-Experts (MoE) architectures, reasoning and non-reasoning modes, quantization variants and parameter scales from 0.6B to 671B. Focusing on three representative black-box single-pass methods, including token probability-based uncertainty (TPU), numerical verbal uncertainty (NVU), and linguistic verbal uncertainty (LVU), we systematically evaluate uncertainty calibration and selective classification using the challenging MMLU-Pro benchmark, which covers both reasoning-intensive and knowledge-based tasks. Our results show that LVU consistently outperforms TPU and NVU, offering stronger calibration and discrimination while being more interpretable. We also find that high accuracy does not imply reliable uncertainty, and that model scale, post-training, reasoning ability and quantization all influence estimation performance. Notably, LLMs exhibit better uncertainty estimates on reasoning tasks than on knowledge-heavy ones, and good calibration does not necessarily translate to effective error ranking. These findings highlight the need for multi-perspective evaluation and position LVU as a practical tool for improving the reliability of LLMs in real-world settings.
\end{abstract}
\begin{figure}[!h]
    \centering    \includegraphics[width=\linewidth]{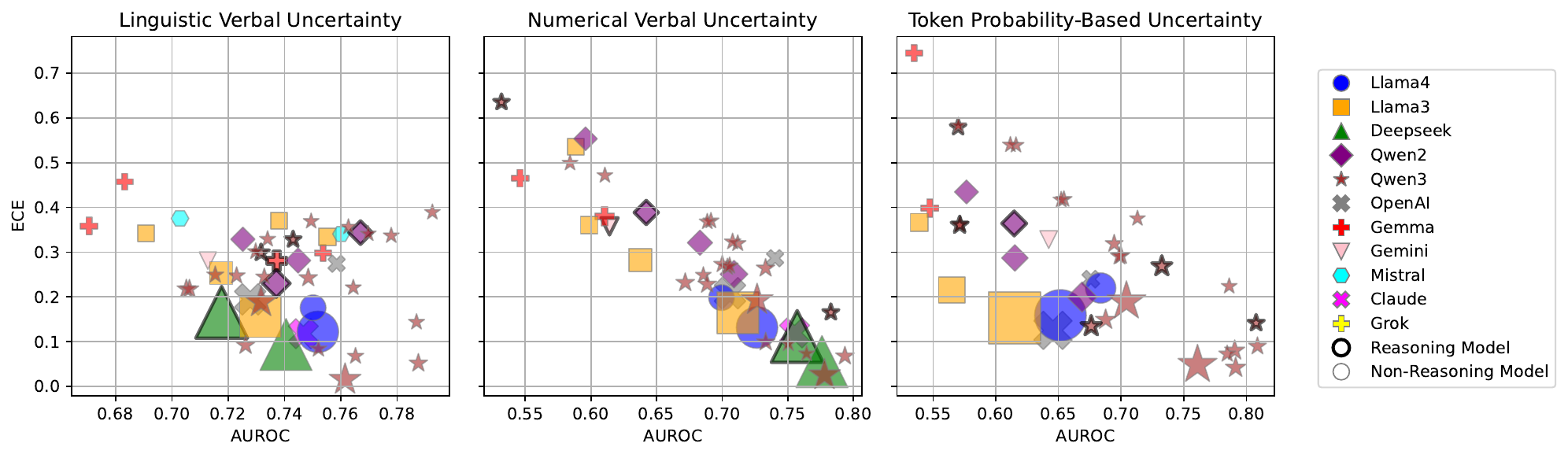}
    \caption{\textbf{AUROC vs. ECE across various LLMs, evaluated using Linguistic Verbal Uncertainty (LVU), Numerical Verbal Uncertainty (NVU) and Token Probability-based Uncertainty (TPU).}~Model families are distinguished by color and marker shape. Reasoning-focused models are highlighted with thick black borders, while non-reasoning models are shown with dashed gray borders. Marker size reflects model size.}
    \label{fig:model_ece_vs_auroc_all_methods}
\end{figure}

\section{Introduction}
As large language models (LLMs) become increasingly integrated into high-stakes applications such as medical decision support and legal consultation~\cite{thirunavukarasu2023large, van2024adapted, abacha2024medec, hadi2023large, lai2024large}, growing concerns have emerged around the public’s over-reliance on their outputs~\cite{passi2022overreliance, isaca2024euai}. For instance, a New York lawyer relied on ChatGPT for legal research and subsequently submitted fabricated case citations generated by the model~\cite{bbc2023debtceiling}. A promising way to mitigate such risks is to equip LLMs with \emph{reliable uncertainty estimation}—enabling the model not only to quantify its uncertainty in a given response (\emph{uncertainty calibration}), but also to distinguish between likely correct and incorrect predictions (\emph{selective classification}). These capabilities support safer deployment: systems can abstain from uncertain predictions, flag high-risk outputs for human review, or adjust decision thresholds dynamically. In short, trustworthy LLMs must not only be accurate, but also well-calibrated and uncertainty-aware.

To extract uncertainty from large language models, a variety of methods have been proposed, which can be broadly categorized into five classes, as summarized in Table~\ref{tab:Uncertainty estimation methods for LLMs.}. (i) \emph{Token Probability-Based Uncertainty (TPU)} estimates uncertainty from output token probabilities, such as perplexity~\cite{duan2023shifting, lambsemantic}; (ii) \emph{Internal state-based methods} leverage model internals like hidden states or attention maps~\cite{kossen2024semantic, ji2025calibrating, zhu2025mitigating}, but require white-/gray-box access. In contrast, black-box-friendly approaches include: (iii) \emph{Consistency-based methods}, which compute uncertainty from the disagreement across multiple generations~\cite{kuhn2023semanticuncertaintylinguisticinvariances, manakul2023selfcheckgpt}; (iv) \emph{Numerical Verbal Uncertainty (NVU)}, where the model self-reports a numeric uncertainty score~\cite{tian2023just, xiong2024llmsexpressuncertaintyempirical}; and (v) \emph{Linguistic Verbal Uncertainty (LVU)}, which infers uncertainty from hedging language in responses, evaluated via a separate judging LLM~\cite{yona2024can, belem2024perceptions, ji2025calibrating}. Besides, LVU offers interpretable and user-friendly uncertainty signals.

Yet, to date there has been no comprehensive study that systematically evaluates uncertainty estimation across the diverse landscape of modern LLMs. \citet{xiong2024llmsexpressuncertaintyempirical} provide an empirical evaluation of NVU, but their study focuses only on prompting techniques. \citet{yona2024can} explore LVU, but evaluate only a small number of models. The most related work, \citet{zhu2023calibration}, analyzes LLM calibration but does not cover newer models or estimation methods. Given the rapid evolution of LLMs~\cite{guo2025deepseek, qwen3blog2024, meta2025llama4, openai2024o3o4mini, xai2024grok3}, the saturation of older benchmarks~\cite{joshi-etal-2017-triviaqa, hendrycks2020measuring}, and the emergence of novel uncertainty estimation techniques~\cite{xiong2024llmsexpressuncertaintyempirical, yona2024can}, there is a clear need for a comprehensive evaluation.

In this work, we present such an evaluation, systematically studying uncertainty estimation in 80 state-of-the-art LLMs. Our benchmark covers both closed-source (e.g., OpenAI GPT, Anthropic Claude, Google Gemini) and open-source (e.g., Meta LLaMA, Qwen3, LLaMA-4, DeepSeek, Mistral) models, ranging from 0.6B to 671B parameters. We assess uncertainty through two complementary tasks: \emph{uncertainty calibration}, which evaluates how well predicted uncertainty aligns with correctness (measured by ECE), and \emph{uncertainty-based selective classification}, which assesses the model’s ability to distinguish correct from incorrect predictions (measured by AUROC). All evaluations are conducted on \textit{MMLU-Pro}~\cite{wang2024mmlu}, a recent and challenging benchmark combining reasoning-intensive (e.g., math, physics) and knowledge-heavy (e.g., law, history) multiple-choice tasks.

We exclude methods that are either computationally expensive or infeasible in black-box settings. Specifically, multi-generation approaches such as semantic entropy~\cite{kuhn2023semanticuncertaintylinguisticinvariances}, which capture question-level uncertainty rather than uncertainty about a specific output, and internal state-based methods that rely on hidden activations or logits unavailable in real-world deployments. Instead, we focus on three black-box, single-pass uncertainty estimation methods: 
(i) \textbf{Token Probability-Based Uncertainty (TPU)}~\cite{lambsemantic}, which estimates uncertainty from token-level probabilities using the inverse of perplexity over the generated response; 
(ii) \textbf{Numerical Verbal Uncertainty (NVU)}~\cite{xiong2024llmsexpressuncertaintyempirical}, where the model is explicitly prompted to produce a self-rated uncertainty score (e.g., between 0 and 100) alongside its answer; and 
(iii) \textbf{Linguistic Verbal Uncertainty (LVU)}~\cite{yona2024can}, which infers uncertainty from hedging expressions (e.g., ``probably'', ``might'') embedded in the response, judged by a separate LLM.

Our large-scale evaluation reveals several key findings about uncertainty estimation in modern LLMs. First, LVU consistently outperforms both TPU and NVU methods, yielding better calibration (lower ECE) and discrimination (higher AUROC) across models and tasks. Second, we find that higher accuracy does not imply better uncertainty estimate—some high-accuracy models (e.g., GPT-4.1) are poorly calibrated, while others (e.g., Qwen3-235B-A22B) achieve reliable uncertainty estimates despite lower accuracy. Third, model design and training factors such as scale, post training, quantization and reasoning specialization significantly affect uncertainty performance. Reasoning model in particular reduces overconfidence problem in high-confidence predictions. We also observe that Mixture-of-Experts (MoE) models may offer advantages over dense models in estimating uncertainty, although this trend requires further validation. Lastly, we find that uncertainty estimation is task-sensitive: LLMs perform more reliably on reasoning-oriented subjects (e.g., math, biology) than on knowledge-seeking ones (e.g., law, history), and good calibration does not necessarily imply good selective classification—highlighting the need for multi-metric evaluation.

Our contributions can be summarized as follows:
\begin{enumerate}
    \item \textbf{Comprehensive evaluation at scale:} We conduct the first comprehensive study of uncertainty estimation across 80 LLMs, covering open- and closed-source models, dense and MoE architectures, and parameter scales ranging from 0.6B to 671B.

    \item \textbf{First large-scale analysis of linguistic uncertainty:} We present the first extensive evaluation of LVU, showing that it consistently outperforms both TPU and NVU in calibration and selective classification. This establishes LVU as a strong and practical approach for future work on uncertainty-aware LLMs.

    \item \textbf{Empirical insights to guide future development:} Our analysis uncovers a range of actionable empirical findings such as the effect of reasoning model, model scale, architecture, quantization and task type on uncertainty estimation, offering valuable guidance for future method design and theoretical analysis.
\end{enumerate}

\begin{table}[]
    \caption{Taxonomy of uncertainty estimation methods for LLMs.}
    \centering
    \scalebox{0.6}{
    \begin{tabular}{llccp{8cm}}
    \toprule
    \textbf{Category} & \textbf{Representative Work} & \textbf{Black-Box} & \textbf{Single Gen.} & \textbf{Description} \\
    \midrule
    \multirow{2}{*}{Token Probability-Based Uncertainty} & \multirow{2}{*}{\cite{duan2023shifting, lambsemantic, burns2022discovering}} & \multirow{2}{*}{\ding{55}} & \multirow{2}{*}{\ding{51}} & Computes uncertainty from token-level probabilities, such as perplexity over the response. \\
    \midrule
    \multirow{2}{*}{Internal State-Based Uncertainty} & \multirow{2}{*}{\cite{kossen2024semantic, ji2025calibrating, kadavath2022language}} & \multirow{2}{*}{\ding{55}} & \multirow{2}{*}{\ding{51}} & Extracts uncertainty from internal activations like hidden states or attention weights. \\
    \midrule
    \multirow{2}{*}{Consistency-Based Uncertainty} & \multirow{2}{*}{\cite{kuhn2023semanticuncertaintylinguisticinvariances, manakul2023selfcheckgpt, farquhar2024detecting}} & \multirow{2}{*}{\ding{51}} & \multirow{2}{*}{\ding{55}} & Estimates uncertainty via disagreement across multiple sampled generations. \\
    \midrule
    \multirow{2}{*}{Numerical Verbal Uncertainty} & \multirow{2}{*}{\cite{xiong2024llmsexpressuncertaintyempirical, tian2023just}} & \multirow{2}{*}{\ding{51}} & \multirow{2}{*}{\ding{51}} & Prompts the model to output a numeric score, e.g., “Uncertainty score: 0.35”. \\
    \midrule
    \multirow{2}{*}{Linguistic Verbal Uncertainty} & \multirow{2}{*}{\cite{yona2024can, belem2024perceptions, ji2025calibrating}} & \multirow{2}{*}{\ding{51}} & \multirow{2}{*}{\ding{51}} & Infers uncertainty from hedging language in the response (e.g., “probably”, “might”), judged by another model. \\
    \bottomrule
    \end{tabular}}
    
    \label{tab:Uncertainty estimation methods for LLMs.}
\end{table}

\section{LLM Uncertainty Estimation and Evaluation}
To estimate uncertainty in large language models (LLMs), a broad range of methods have been proposed. These can be grouped into five major categories, as summarized in Table~\ref{tab:Uncertainty estimation methods for LLMs.}:  
(i) \emph{Token Probability-Based Uncertainty (TPU)} estimates uncertainty from output token probabilities (e.g., perplexity)~\cite{duan2023shifting, lambsemantic}, aligning closely with the autoregressive nature of LLMs;  
(ii) \emph{Internal state-based methods} leverage hidden states or attention weights~\cite{kossen2024semantic, ji2025calibrating, zhu2025mitigating}, but require white- or gray-box access, limiting their applicability in practice.

Three black-box-friendly alternatives have gained increasing attention:  
(iii) \emph{Consistency-based methods} compute uncertainty based on disagreement across multiple sampled generations~\cite{kuhn2023semanticuncertaintylinguisticinvariances, manakul2023selfcheckgpt}, capturing question-level uncertainty but at a high computational cost;  
(iv) \emph{Numerical Verbal Uncertainty (NVU)} prompts the model to output a self-reported scalar (e.g., “uncertainty: 0.35”)~\cite{tian2023just, xiong2024llmsexpressuncertaintyempirical};  
(v) \emph{Linguistic Verbal Uncertainty (LVU)} infers uncertainty from hedging expressions (e.g., “probably”, “might”) embedded in the response, evaluated using a separate judging LLM~\cite{yona2024can, belem2024perceptions, ji2025calibrating}, which offer more natural and interpretable uncertainty signals for human-facing applications.

\subsection{LLM Uncertainty Estimation}
In this work, we define \emph{uncertainty} as a scalar in \([0, 1]\), where 0 indicates complete certainty and 1 indicates maximum uncertainty. Formally, we write \( u_\phi(\mathbf{y} \mid \mathbf{x}) = 1 - p_\phi(\mathbf{y} \mid \mathbf{x}) \), where \( \mathbf{x} \) is the input, \( \mathbf{y} \) is the generated output, and \( p_\phi \) is the model’s confidence. Our study focuses on \emph{pointwise uncertainty}—the model’s uncertainty about its actual response in a single generation—rather than question-level uncertainty aggregated across multiple outputs. Thus, we adopt TPU, NVU, and LVU as our baseline methods for systematic evaluation.

\paragraph{Token Probability-Based Uncertainty (TPU)}
\label{sec:token-uncertainty}
This method estimates uncertainty using the model’s output token probabilities. Let $\mathcal{V}$ denote the vocabulary, and $\mathbf{x} \in \mathcal{V}^l$ be an input prompt of $l$ tokens. The model generates a response $\mathbf{y} = (y_1, \ldots, y_n) \in \mathcal{V}^n$, where 
$n$ is the number of tokens in the generated response. The autoregressive log-likelihood is given by:
\begin{equation}
\log p_\phi(\mathbf{y} \mid \mathbf{x}) = \sum_{i=1}^{n} \log p_\phi(y_i \mid \mathbf{y}_{<i}, \mathbf{x}), \quad \text{where } \mathbf{y}_{<i} = (y_1, \ldots, y_{i-1}).
\end{equation}

To normalize for sequence length, we compute the average log-likelihood:
\begin{equation}
\bar{\ell}(\mathbf{y} \mid \mathbf{x}) = \frac{1}{n} \sum_{i=1}^{n} \log p_\phi(y_i \mid \mathbf{y}_{<i}, \mathbf{x}).
\end{equation}

Following~\cite{lambsemantic}, we define the \emph{Token Probability-Based Uncertainty} as the complement of the exponentiated average log-likelihood of the generated response:
\begin{equation}
u_\phi^{\text{TPU}}(\mathbf{y} \mid \mathbf{x}) := 1 - \exp\left( \bar{\ell}(\mathbf{y} \mid \mathbf{x}) \right) = 1 - p_\phi(\mathbf{y} \mid \mathbf{x})^{1/n} \in [0, 1],
\end{equation}
where lower values indicate greater confidence. This method is broadly applicable—even to gray-box models—and serves as a strong baseline for uncertainty estimation.

\paragraph{Numerical Verbal Uncertainty (NVU)}
NVU elicits explicit uncertainty estimates directly from the LLM. The model is prompted to output a numeric uncertainty score alongside its response, offering an interpretable form of subjective uncertainty. We incorporate chain-of-thought (CoT) prompting to extract better uncertainty estimation following~\cite{xiong2024llmsexpressuncertaintyempirical}. The complete NVU prompting templates are provided in Appendix~\ref{appendix:Prompts and Regular Expressions}.

\paragraph{Linguistic Verbal Uncertainty (LVU)}
In contrast to NVU, LVU captures uncertainty implicitly through hedging language embedded in the model’s response—e.g., phrases like ``probably'', ``might'', or ``possibly''. Rather than providing a numeric score, the model expresses its uncertainty through natural language, aligning with how humans convey uncertainty in conversation. To quantify these implicit signals, we adopt the approach of~\cite{yona2024can}, using \texttt{LLaMA-4-Maverick-17B-128E-Instruct} as a separate evaluator to interpret and score linguistic uncertainty.  This decouples generation from evaluation: the primary model produces a response, and an external model assesses the level of uncertainty conveyed linguistically. We provide an empirical analysis of the LLM judge’s effectiveness in the Appendix~\ref{appendix:llmasajudge}, along with details of the generation and evaluation prompt templates in Appendix~\ref{appendix:Prompts and Regular Expressions}.


\subsection{Evaluation of Uncertainty Estimation}

To assess the quality of uncertainty estimation in large language models (LLMs), we consider two widely adopted evaluation tasks: \textit{uncertainty calibration} and \textit{uncertainty-based selective classification}. \textbf{Uncertainty calibration} measures how well predicted uncertainty values reflect the true likelihood of correctness, typically using the \textit{Expected Calibration Error} (ECE)~\cite{guo2017calibration}. In contrast, \textbf{selective classification} evaluates how well uncertainty scores distinguish correct from incorrect predictions, using the \textit{Area Under the Receiver Operating Characteristic curve} (AUROC)~\cite{mucsanyi2024benchmarking}.

\paragraph{Uncertainty Calibration} 
Uncertainty calibration evaluates whether a model’s predicted uncertainty values correspond to actual correctness likelihoods. For example, among predictions with uncertainty 0.2 (i.e., 0.8 confidence), approximately 80\% should be correct for a well-calibrated model. Formally, perfect calibration satisfies:
\begin{equation}
P(\hat{Y} = Y \mid U = u) = 1 - u, \quad \forall u \in [0, 1],
\end{equation}
where $\hat{Y}$ is the model's prediction, $Y$ is the ground truth, and $U$ is the predicted uncertainty. In practice, calibration is quantified using \textit{Expected Calibration Error (ECE)}, which partitions predictions into $M$ uncertainty bins and computes the weighted average absolute difference between empirical accuracy and the implied certainty $(1 - \text{uncertainty})$:
\begin{equation}
\text{ECE} = \sum_{m=1}^{M} \frac{|B_m|}{n} \left| \text{acc}(B_m) - (1 - \text{unc}(B_m)) \right|,
\end{equation}
where $|B_m|$ is the number of predictions in bin $m$, $n$ is the total number of predictions, and $\text{unc}(B_m)$ is the average uncertainty in bin $m$. Lower ECE indicates better calibration. We adopt number of bins M = 10 for evaluation. We also use \textit{adaptive ECE}~\cite{nixon2019measuring} to ensure robustness against skewed uncertainty distributions and reported in the Appendix~\ref{appendix:adaptiveece}. Reliability diagrams~\cite{niculescu2005predicting} are also used for qualitative assessment.

\paragraph{Selective Classification}
Selective classification assesses the discriminative power of uncertainty scores in ranking correct versus incorrect predictions. The most common metric is AUROC, which measures the probability that a randomly chosen correct prediction is assigned lower uncertainty than a randomly chosen incorrect one. An AUROC of 0.5 corresponds to random guessing, while a value closer to 1.0 indicates stronger discrimination.

\paragraph{Models}
We evaluate 80 state-of-the-art LLMs, including both open- and closed-source models, with parameter scales ranging from 0.6B to 671B. Representative models include: \textbf{Open-source}—Meta LLaMA 3 (1B–405B)~\cite{grattafiori2024llama}, LLaMA 4 (109B, 400B)~\cite{meta2025llama4}, DeepSeek~\cite{guo2025deepseek, liu2024deepseek}, Qwen2.5/Qwen3 (0.6B–235B)~\cite{yang2024qwen2, qwen3blog2024}, Mistral~\cite{jiang2024identifying}, and Google Gemma~\cite{team2024gemma}; \textbf{Closed-source}—OpenAI GPT-4.1/GPT-4o/o3/o4~\cite{openai2024gpt4o, openai2024gpt41, openai2024o3o4mini}, Anthropic Claude 3~\cite{anthropic2024claude3}, Google Gemini~\cite{team2024gemini}, and xAI Grok~\cite{xai2024grok3}. A complete model list is in Appendix~\ref{appendix:evaluation results}.

\paragraph{Dataset}
We conduct all evaluations on MMLU-Pro~\cite{wang2024mmlu}, a comprehensive benchmark released in June 2024 that includes both reasoning-centric tasks (e.g., math, physics, engineering) and knowledge-centric tasks (e.g., law, history, psychology). Compared to prior benchmarks, MMLU-Pro is more challenging and less saturated, making it a robust testbed for evaluating uncertainty estimation.

\paragraph{Prompt Design}
To enable consistent and fair uncertainty extraction across methods, we adopt a unified prompt template that combines numerical verbal uncertainty and linguistic verbal uncertainty formats~\cite{xiong2024llmsexpressuncertaintyempirical, yona2024can}, with an additional concise CoT component. This design allows fair comparison across all models and uncertainty modalities. Detailed prompt formats are provided in Appendix~\ref{appendix:Prompts and Regular Expressions}.

\paragraph{Uncertainty Extraction}
All models are prompted to produce answers in a predefined format. We use regular expressions to extract uncertainty scores. Outputs that deviate from the expected formats are excluded from analysis. Extraction scripts and format rules are described in Appendix~\ref{appendix:Prompts and Regular Expressions}.

\section{Results and Discussion}
\subsection{Larger Models Generally Yield More Reliable Uncertainty Estimates}

\begin{figure}
    \centering
    \includegraphics[width=\linewidth]{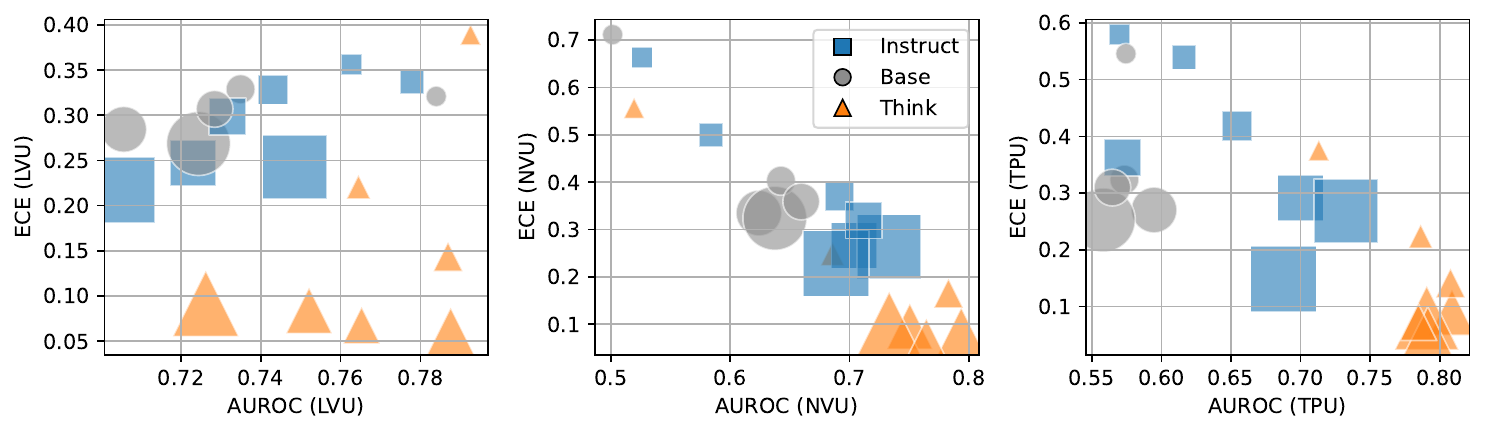}
    \caption{Uncertainty estimation performance of Qwen3 across three variants: no post-training (Base), with post-training (Intruct), and reasoning-enhanced (Think). Point size reflects model scale.}
    \label{fig:Comparison among Qwen3 Models}
\end{figure}

Model scale plays a critical role in shaping uncertainty estimation. While earlier studies in conventional deep learning suggest that larger models tend to be more overconfident and poorly calibrated~\cite{guo2017calibration}, our findings in the LLM setting suggest otherwise. Across all three methods, we observe that larger models tend to produce more calibrated and discriminative uncertainty scores. This trend is evident in Figures~\ref{fig:model_ece_vs_auroc_all_methods}, especially under NVU and TPU.

To isolate the effect of model size, we examine the Qwen3 series trained under similar conditions, ranging from 0.6B to 32B parameters. As shown in Figure~\ref{fig:Comparison among Qwen3 Models}, larger models such as Qwen3-14B and Qwen3-32B consistently outperform their smaller counterparts in both calibration (lower ECE) and selective classification (higher AUROC). However, we also find that performance gains saturate at scale, and that mid-sized models like Qwen3-8B can achieve competitive results. In contrast, lightweight models such as Qwen3-0.6B and Qwen3-1.7B struggle to produce meaningful uncertainty scores, often approaching random guessing. This underscores the practical limitations of compact LLMs in scenarios that demand reliable uncertainty quantification.

Interestingly, under LVU, this trend appears partially reversed when evaluating AUROC. Upon further inspection, we find that smaller models tend to generate responses that poorly follow instructions or omit uncertainty cues altogether. The judge model tends to assign high LVU uncertainty to such non-compliant outputs, many of which are indeed incorrect. This results in strong discriminative uncertainty for smaller models—not because their uncertainty estimates are intrinsically better, but because instruction violations correlate strongly with failure.


\subsection{Post-Training Enhances Uncertainty Estimation}

Modern post-training techniques, including instruction fine-tuning and Direct Preference Optimization (DPO), are designed to align LLM outputs with human expectations and communication styles. Our results show that these techniques also significantly enhance uncertainty estimation. By comparing Qwen3-Base (no post-training) and Qwen3-Instruct (with post-training) models as shown in Figure~\ref{fig:Comparison among Qwen3 Models}, we observe consistent improvements across all three methods: post-trained models achieve lower ECE and higher AUROC, indicating both better calibration and stronger ability to distinguish between correct and incorrect predictions. This improvement is robust across model sizes. These findings suggest that post-training helps models not only generate more helpful outputs but also express uncertainty more reliably. For all other model series, we include only post-trained variants in our main evaluation to ensure fair comparisons. This aligns with and extends the findings of~\cite{zhu2023calibration}, and highlights post-training as an effective mechanism for enhancing LLM uncertainty awareness.

\begin{figure}[t]
    \centering
    \begin{subfigure}[b]{0.24\linewidth}
        \includegraphics[width=\linewidth]{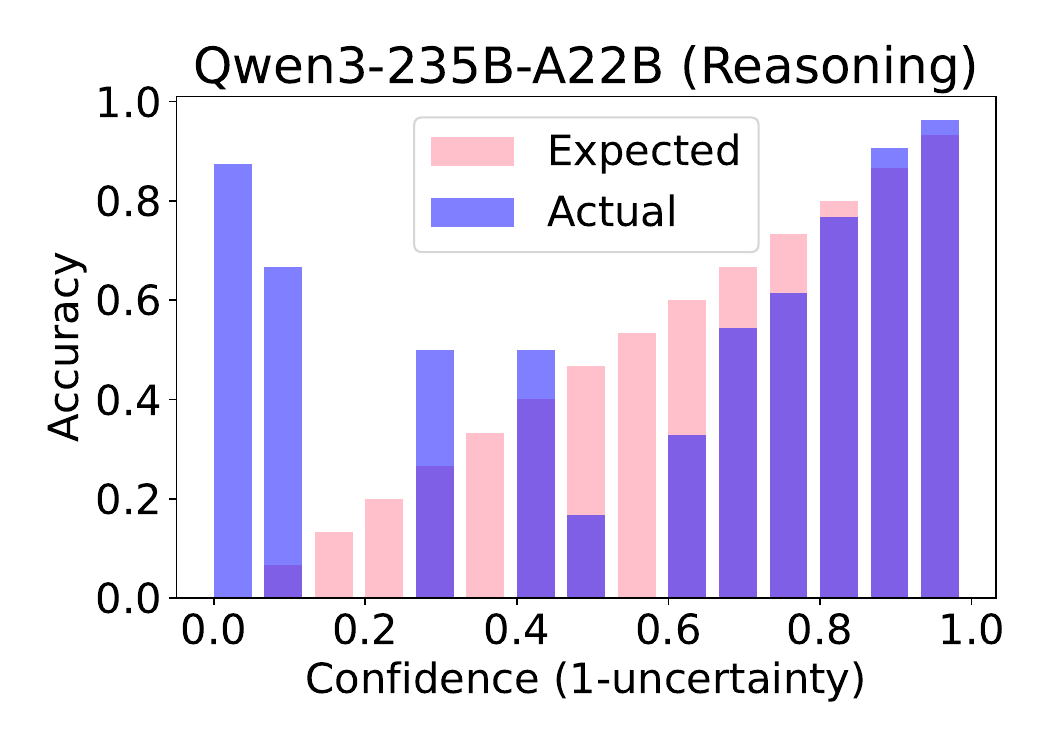}
    \end{subfigure}
    \begin{subfigure}[b]{0.24\linewidth}
        \includegraphics[width=\linewidth]{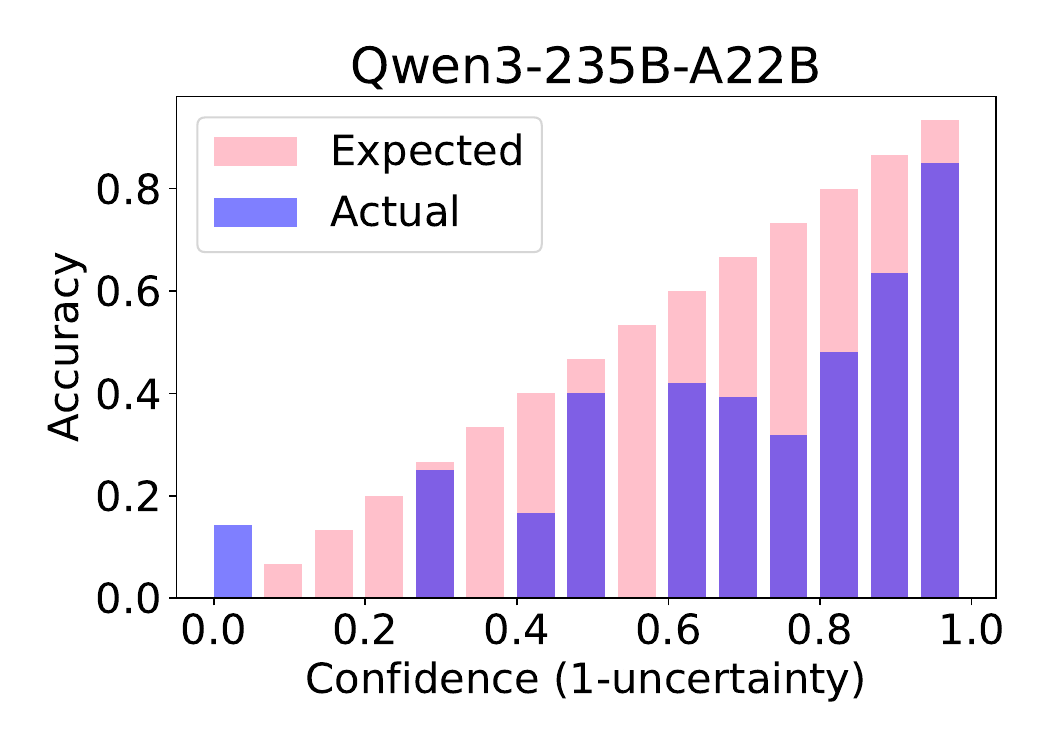}
    \end{subfigure}
    \begin{subfigure}[b]{0.24\linewidth}
        \includegraphics[width=\linewidth]{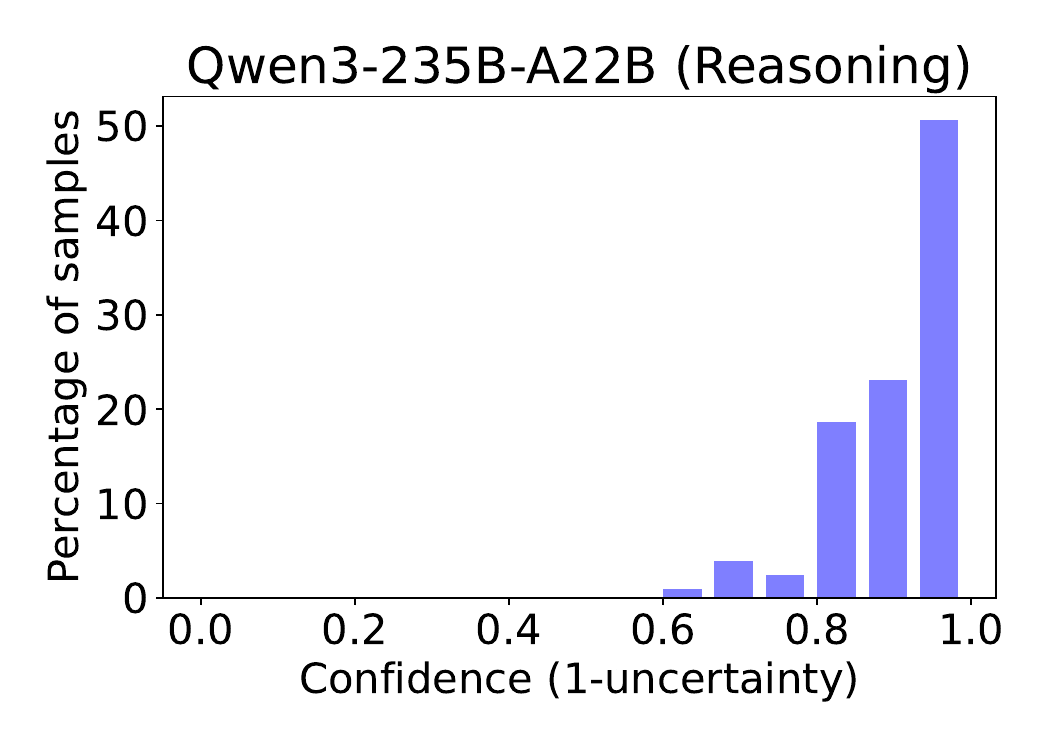}
    \end{subfigure}
    \begin{subfigure}[b]{0.24\linewidth}
        \includegraphics[width=\linewidth]{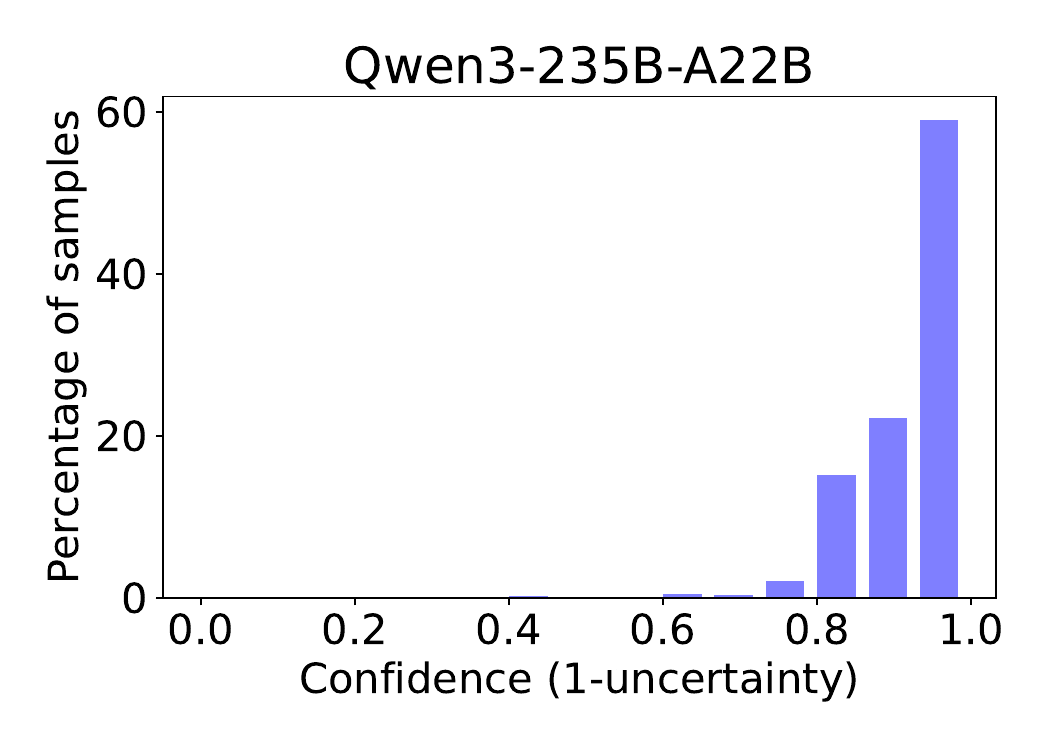}
    \end{subfigure}
    \caption{Reliability diagram and bin density between reasoning and non-reasoning LLMs.}
    \label{fig:Reliability Diagram and Bin Density between Reasoning and non Reasoing LLM}
\end{figure}

\subsection{Reasoning Mode Improves Uncertainty Without Needing Higher Accuracy}
Qwen3 support both standard and reasoning-enhanced modes (denoted as `Instruct' and `Think'). Comparing these variants across model sizes (Figure~\ref{fig:Comparison among Qwen3 Models}), we find that reasoning mode consistently improves uncertainty estimation. For example, Qwen3-4B in reasoning mode matches or exceeds the uncertainty performance of much larger models like GPT-4.1, as shown in Figures \ref{fig:model_ece_vs_auroc_all_methods}. Figure~\ref{fig:reasoning_comparison_and_moe} (left) shows reasoning-enhanced models have marked gains in calibration and modest improvements in selective classification, compared to non-reasoning models (averaged over all evaluated models).

To understand this effect in depth, we analyze reliability diagrams and bin density plots for representative models (Figure~\ref{fig:Reliability Diagram and Bin Density between Reasoning and non Reasoing LLM}). Reasoning variants exhibit better alignment between predicted and actual correctness, particularly in high-confidence regions. Moreover, reasoning reduces the proportion of highly overconfident predictions by more than 20\%, indicating its effectiveness in curbing model overconfidence. These results demonstrate that reasoning models not only enhances interpretability but also significantly boosts the model’s ability to estimate uncertainty.

\subsection{MoE Models May Offer Improved Uncertainty Estimation over Dense Models}
Mixture-of-Experts (MoE) has emerged as a compelling architectural paradigm for scaling LLMs efficiently, with recent adoption in models such as Qwen3-MoE, LLaMA-4, Grok-3, and DeepSeek. Unlike fully dense models, MoE models activate only a subset of parameters per input, potentially enabling more adaptive and specialized behavior. However, due to their recent emergence, head-to-head comparisons with dense counterparts remain limited.

One relatively fair comparison is between Qwen3-30B-A3B (MoE) and Qwen3-32B (dense), which share similar total parameter counts and follow closely matched training protocols~\cite{qwen3report2024}. The Qwen3-30B-A3B model activates 3B parameters per forward pass, while Qwen3-32B is fully dense.

\begin{figure}
    \centering
    \begin{subfigure}[b]{0.49\linewidth}
        \includegraphics[width=\linewidth]{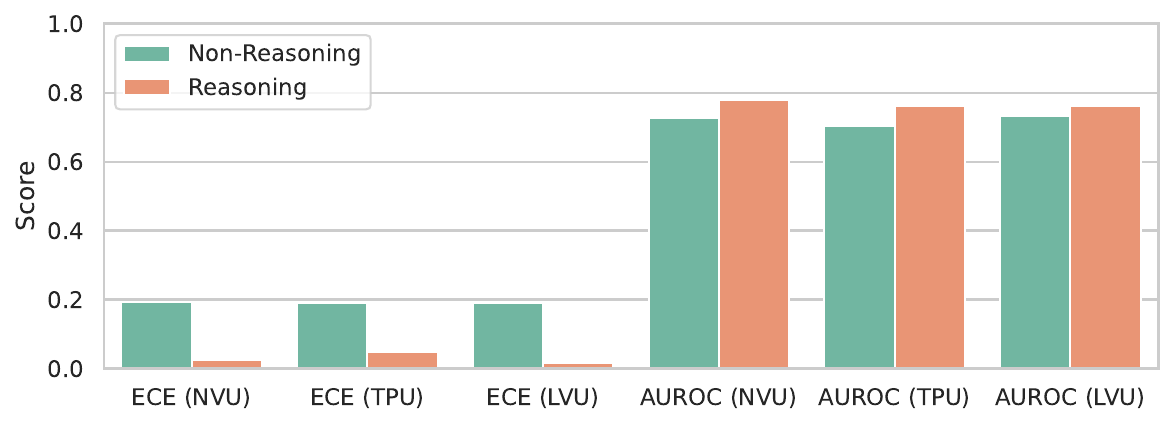}
    \end{subfigure}
    \begin{subfigure}[b]{0.49\linewidth}
        \includegraphics[width=\linewidth]{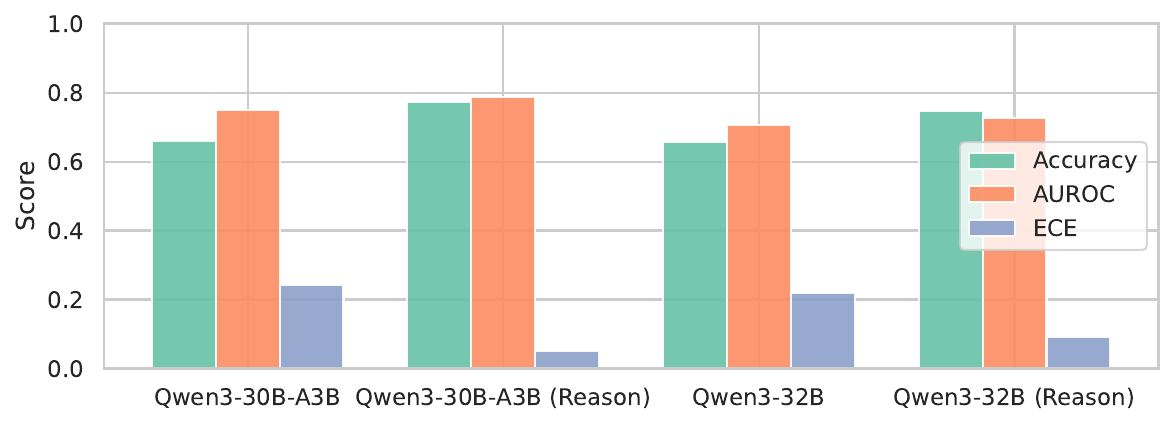}
    \end{subfigure}
    \caption{\textbf{Left:} Comparison of Reasoning vs. Non-Reasoning on ECE and AUROC. Results are averaged over all reasoning/non-reasoning models. \textbf{Right:} Comparison between MoE and dense models within the Qwen3 series under LVU.}
    \label{fig:reasoning_comparison_and_moe}
\end{figure}

As shown in Figure~\ref{fig:reasoning_comparison_and_moe} (right), Qwen3-30B-A3B consistently outperforms its dense counterpart in both reasoning and non-reasoning tasks when evaluated using LVU, achieving better calibration and more reliable selective classification. This result suggests that the conditional computation of MoE may facilitate more nuanced uncertainty estimation. Nevertheless, the sample size of comparable MoE–dense model pairs remains small, and further systematic comparisons are needed to confirm whether MoE architectures generally confer an advantage in uncertainty estimation.

\subsection{High Accuracy Does Not Guarantee Reliable Uncertainty Estimation}
\label{main:High Accuracy Does Not Guarantee Reliable Uncertainty Estimation}


While recent advancements in LLM development have predominantly emphasized improving accuracy, our results reveal that high accuracy does not necessarily imply reliable uncertainty estimation under LVU. As shown in Figure~\ref{fig:higheracc_lowuncertainty_and_comparison_of_confidence_extraction_methods} left, several top-performing models—including LLaMA-4-Maverick, LLaMA-3.1-405B, and GPT-4.1—exhibit strong accuracy but poor uncertainty estimation under TPU, as indicated by elevated ECE and reduced AUROC scores. 
We observe similar discrepancies across the NVU and LVU methods, detailed in the Appendix~\ref{appendix:High Accuracy Does Not Guarantee Reliable Uncertainty Estimation}. 

Notably, \texttt{Qwen3-235B-A22B (Reasoning)}, which achieves only moderate accuracy (67\%)—about 15 points lower than models such as \texttt{DeepSeek-R1}, \texttt{GPT-4.1}, or \texttt{Grok-3}—yet consistently delivers top-tier uncertainty performance across all three methods. This suggests that reliable uncertainty estimation is not merely a byproduct of better accuracy, but a distinct capability shaped by model architecture, training, and prompting.

\subsection{Linguistic Verbal Uncertainty Shows High Potential for Reliable Uncertainty Estimation}

\begin{figure}
    \centering
    \begin{subfigure}[b]{0.65\linewidth}
        \includegraphics[width=\linewidth]{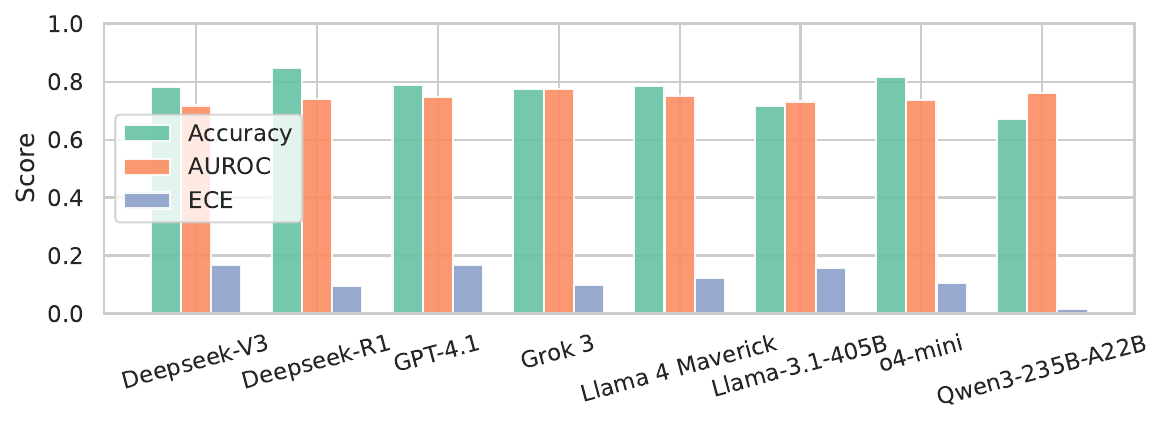}
    \end{subfigure}
    \begin{subfigure}[b]{0.33\linewidth}
        \includegraphics[width=\linewidth]{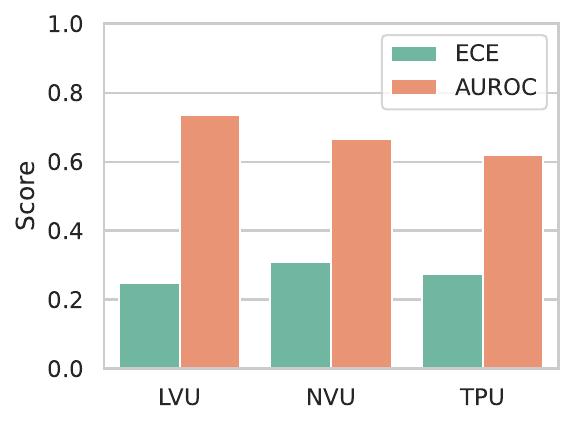}
    \end{subfigure}
    \caption{\textbf{Left:} Comparison of top-tier models under LVU. \textbf{Right:} Comparison of different uncertainty estimation methods.}
    \label{fig:higheracc_lowuncertainty_and_comparison_of_confidence_extraction_methods}
\end{figure}

To compare the overall effectiveness of different uncertainty estimation methods, we compute average AUROC and ECE scores across all 80 evaluated models. As shown in Figure~\ref{fig:higheracc_lowuncertainty_and_comparison_of_confidence_extraction_methods} right, LVU consistently outperforms both NVU and TPU, with roughly 10\% improvement in both AUROC and ECE over the second-best methods. Among the three black-box, single-pass methods, LVU not only achieves the strongest empirical performance, but also offers the most interpretable and human-aligned uncertainty signals, leveraging hedging language such as ``probably'' and ``might.'' Despite these advantages, LVU remains underexplored in the literature, underscoring its potential as a promising direction for future uncertainty-aware LLM development.

\begin{figure}
    \centering
    \begin{subfigure}[b]{0.49\linewidth}
        \includegraphics[width=\linewidth]{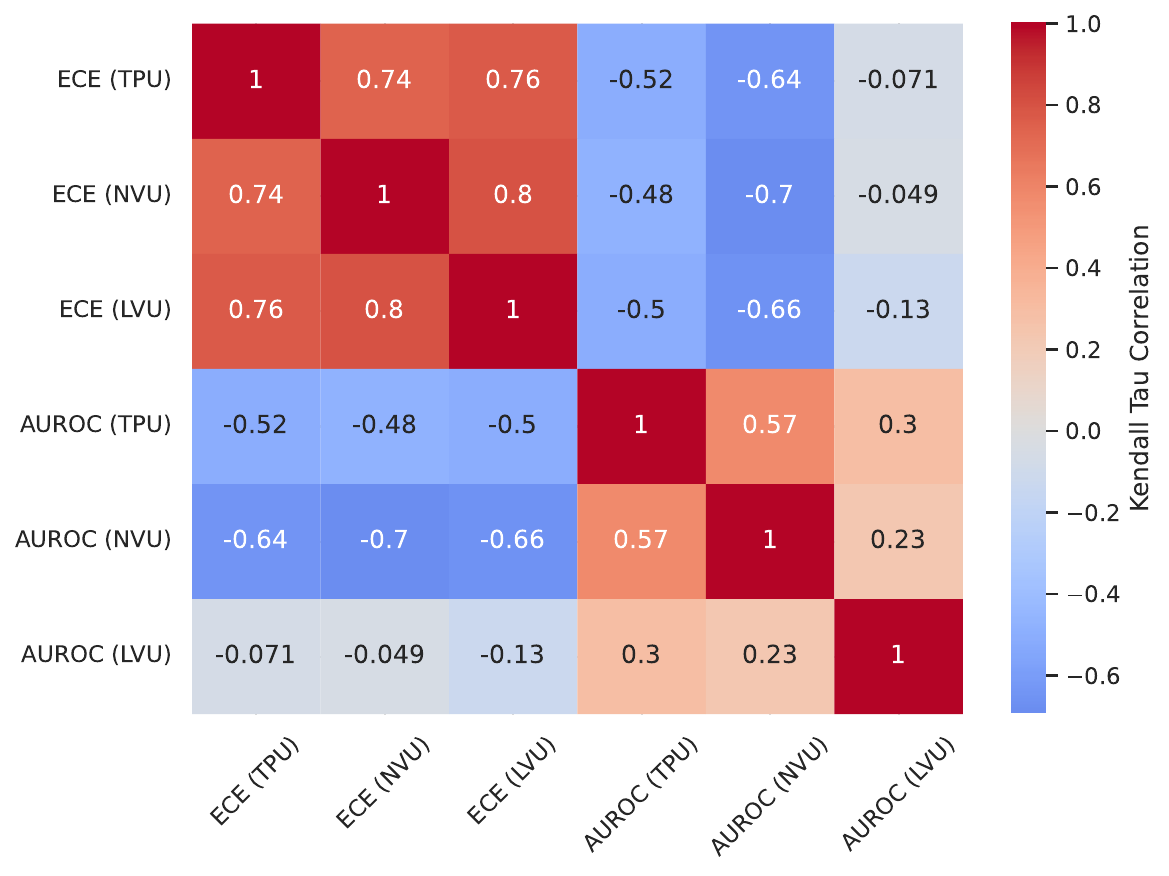}
    \end{subfigure}
    \begin{subfigure}[b]{0.49\linewidth}
        \includegraphics[width=\linewidth]{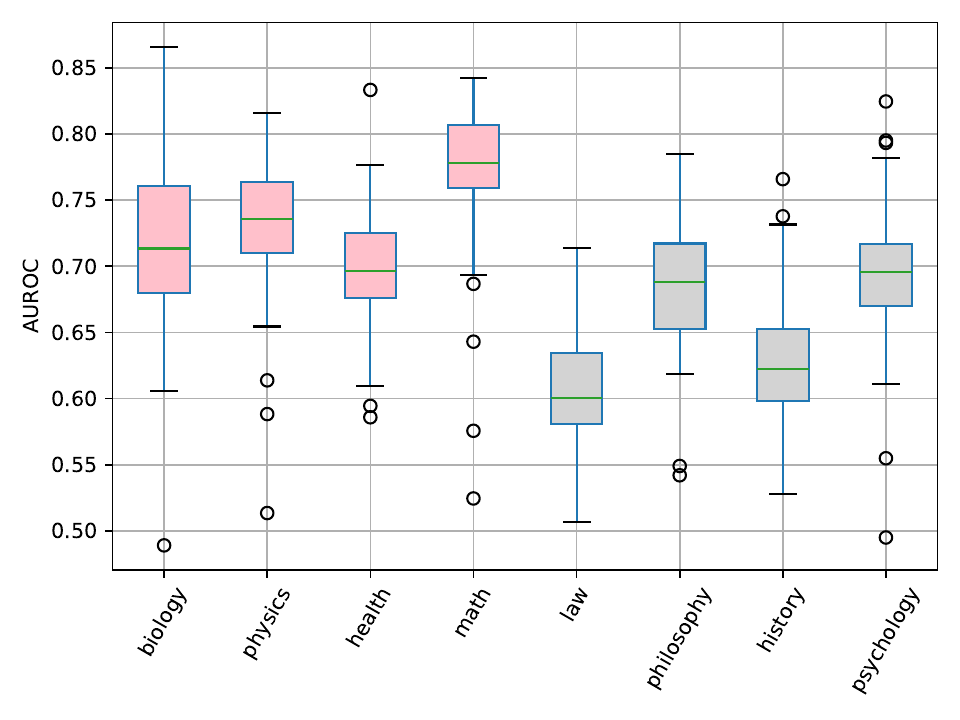}
    \end{subfigure}
    \caption{\textbf{Left:} Kendall Rank Correlation among Uncertainty Estimation Methods. \textbf{Right:} LLMs exhibit better uncertainty estimates in reasoning tasks, as measured by AUROC using LVU.}
    \label{fig:Correlation_between_Uncertainty Estimation_Methods_and_task_difference}
\end{figure}

To better understand their relationship, we compute Kendall rank correlations~\cite{abdi2007kendall} between ECE and AUROC across the three uncertainty estimation methods. As shown in Figure~\ref{fig:Correlation_between_Uncertainty Estimation_Methods_and_task_difference} left, we observe high internal consistency within each metric—particularly between LVU and NVU on ECE—indicating agreement in how well each method captures calibrated uncertainty. However, the correlation between AUROC values across methods is notably weaker, suggesting that different approaches capture distinct signals relevant to ranking performance. We also include the comparisons with multi-generation based uncertainty~\cite{kuhn2023semanticuncertaintylinguisticinvariances, farquhar2024detecting, manakul2023selfcheckgpt} in the Appendix~\ref{appendix:semanticuncertainty} for a comprehensive understanding.


\subsection{Calibration and Selective Classification Capture Distinct Dimensions of Uncertainty}

While both calibration and selective classification are standard tasks for evaluating uncertainty, they assess fundamentally different aspects. Calibration (measured by ECE) evaluates the alignment between predicted uncertainty and actual correctness, while selective classification (measured by AUROC) evaluates how well uncertainty separates correct from incorrect predictions.

Notably, when we check the correlation between metrics in Figure~\ref{fig:Correlation_between_Uncertainty Estimation_Methods_and_task_difference} left, the correlation between calibration (ECE) and selective classification (AUROC) is generally low or even close to zero, especially for LVU, where calibration and ranking scores are nearly uncorrelated. This suggests that good calibration does not imply strong discriminative ability, and vice versa. These findings reinforce the need for multi-metric evaluation in uncertainty estimation, as each metric provides a complementary view of uncertainty reliability.

\subsection{LLMs Estimate Uncertainty More Reliably on Reasoning Tasks}


We further investigate how uncertainty estimation varies across task types. MMLU-Pro offers a diverse set of subjects, covering both reasoning-intensive domains (e.g., math, physics) and knowledge-seeking ones (e.g., law, history). To compare model performance across these categories, we report AUROC scores for 80 models using LVU, as shown in Figure~\ref{fig:Correlation_between_Uncertainty Estimation_Methods_and_task_difference} right, where reasoning-related subjects are marked in red and knowledge-heavy subjects in light gray.

We group reasoning-intensive tasks as \emph{math, biology, health}, and \emph{physics}, and knowledge-seeking tasks as \emph{law, philosophy, history}, and \emph{psychology}. Our analysis reveals that models estimate uncertainty more reliably on reasoning tasks, with an average AUROC improvement of over 10\% compared to knowledge-based tasks. This pattern is robust across all three uncertainty estimation methods, suggesting that the structured nature of reasoning problems may allow LLMs to form more consistent and discriminative uncertainty estimates.

\subsection{Quantization Slightly Degrades Uncertainty Estimation Performance}

\begin{figure}
    \centering
    \includegraphics[width=\linewidth]{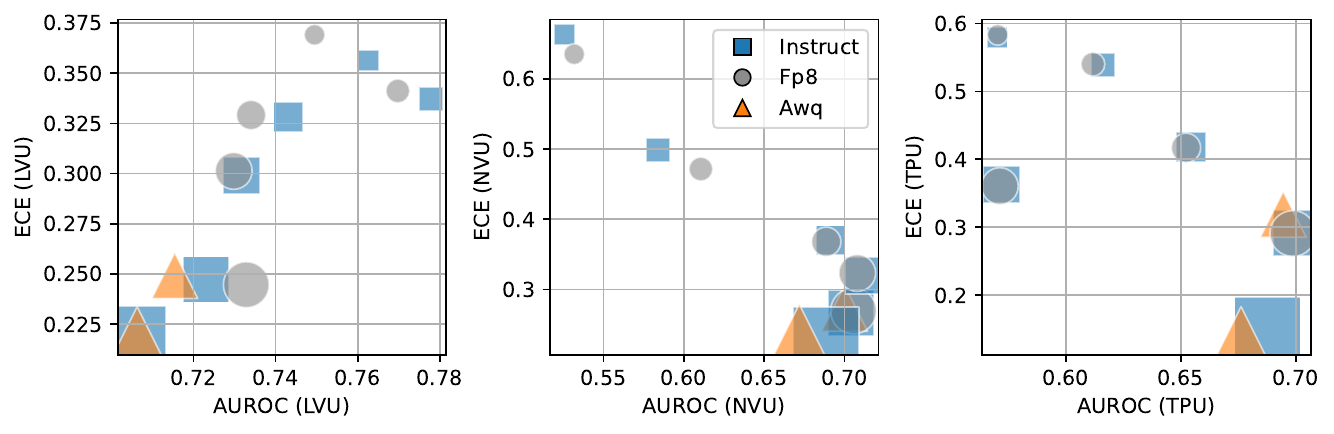}
    \caption{Uncertainty estimation performance of Qwen3 across quantization variants: no quantization (Instruct), FP8, and Activation-aware Weight Quantization (Awq). Point size reflects model scale.}
    \label{fig:qwen3_quantization_effects}
\end{figure}

Quantization has become increasingly important for deploying LLMs on edge and resource-constrained environments. To evaluate its impact on uncertainty estimation, we analyze quantized variants (FP8 and AWQ) of the Qwen3 series. As shown in Figure~\ref{fig:qwen3_quantization_effects}, quantization generally leads to mild degradation in uncertainty estimation, with effects more noticeable in smaller models. Under LVU and TPU, both AUROC and ECE worsen slightly after quantization. Interestingly, NVU shows marginal improvements in small models but degrades as scale increases. Overall, the impact of quantization remains modest, typically within a 5\% margin.

\section{Conclusion and Future Work}

Our large-scale evaluation yields several key insights into understanding and improving uncertainty estimation in LLMs. First, we find that model characteristics—such as size, reasoning specialization, quantization and post-training strategies—significantly influence uncertainty behavior. Second, although underexplored, LVU consistently demonstrates strong empirical performance, indicating that natural language cues can serve as effective and intuitive signals for uncertainty. Third, we observe that predictive accuracy is not a reliable proxy for uncertainty quality: high-accuracy models may still exhibit poor calibration or fail to distinguish correct from incorrect predictions.

These findings highlight several promising directions for future research. First, while our study establishes LVU as a strong and interpretable method, its underlying mechanisms and potential applications remain underexplored. Recent work~\cite{kim2024m} shows that natural-language uncertainty cues can mitigate user over-reliance by encouraging further consideration on uncertain responses. Future studies could investigate how LVU may be leveraged to improve model-user interaction, particularly in high-stakes scenarios. Furthermore, unlike numerical scores or internal logits, LVU aligns more naturally with human communication norms. This makes it a promising candidate for integration into the post-training phase—e.g., as part of preference optimization or instruction tuning—to enhance a model’s \emph{uncertainty awareness} and communicative grounding.

Second, our analysis reveals a structural mismatch between calibration and discrimination: models with low \emph{ECE} do not necessarily achieve high \emph{AUROC}, and vice versa. This suggests the need for new evaluation metrics or learning objectives that jointly reflect both calibration and selective prediction performance. Finally, we advocate for moving beyond static evaluation and incorporating uncertainty estimation into downstream LLM workflows, such as selective answering, fallback prompting, and user-facing warnings. These uncertainty-informed behaviors offer a practical pathway to improving the safety, reliability, and accountability of language models in real-world deployments.

Finally, a limitation of our study lies in the evaluation of LVU: the effectiveness of the LLM judge was only assessed in a small-scale experiment. Its general reliability across tasks, domains, and model families remains unclear. We encourage future work to develop standardized evaluation protocols for assessing the quality and robustness of LLM-based uncertainty judges.

\clearpage
\bibliographystyle{unsrtnat}
\bibliography{neuips_2025}
\clearpage


\appendix
\input{appendix}
\end{document}

%% file: appendix.tex
\section{Related Works}
\label{appendix:relatedwork}
\subsection{More Uncertainty Estimation Approaches}

\paragraph{Multi-Generation Consistency-Based Methods}
Recent advancements in uncertainty estimation for large language models have introduced several methods leveraging multi-generation consistency. SelfCheckGPT~\cite{manakul2023selfcheckgpt} identifies hallucinations by generating multiple responses to the same prompt and assessing their consistency using metrics like BERTScore, n-gram overlap, and natural language inference (NLI). Inconsistencies among responses suggest potential hallucinations. Generating with Confidence~\cite{lin2023generating} proposes evaluating the variability among multiple generated outputs to estimate uncertainty. Semantic Entropy~\cite{kuhn2023semanticuncertaintylinguisticinvariances, farquhar2024detecting} addresses the challenge of semantic equivalence in natural language by clustering semantically similar responses and computing entropy over these clusters, providing a more accurate measure of uncertainty for open-end questions.

\paragraph{Internal State-Based Uncertainty}

Recent research has also explored leveraging the internal states of LLMs to estimate uncertainty and detect hallucinations. ~\citet{azaria2023internal} demonstrate that hidden layer activations can be utilized to predict the truthfulness of generated statements by training a classifier on these activations. ~\citet{li2023inference} introduce Inference-Time Intervention (ITI), a technique that adjusts model activations during inference to enhance truthfulness, significantly improving performance on benchmarks like TruthfulQA. ~\citet{kossen2024semantic} propose Semantic Entropy Probes (SEPs), which estimate semantic uncertainty directly from hidden states without requiring multiple generations, offering a computationally efficient method for hallucination detection.~\citet{ji2025calibrating} identify a linear feature within the model's representation space that governs verbal uncertainty, enabling calibration to reduce overconfident hallucinations.~\citet{burns2022discovering} present an unsupervised approach to uncover latent knowledge by identifying directions in activation space that correspond to logical consistency, allowing accurate yes-no question answering without labeled data. Lastly,~\citet{liu2024uncertainty} propose a supervised method that leverages labeled datasets to estimate uncertainty from hidden activations, demonstrating robust performance across various tasks and model access levels.

\subsection{Study on Uncertainty Estimation for LLMs}

Recent research has delved into the challenges and methodologies of uncertainty estimation in LLMs.~\citet{geng2023survey} provide a comprehensive survey on confidence estimation and calibration techniques.~\citet{huang2024survey} offer a hierarchical categorization of uncertainty estimation methods and drawing relations with traditional machine learning approaches.~\citet{zhu2023calibration} examine token probability-based uncertainty, analyzing the effects of model size and instruction tuning, though they focus on a single estimation method and do not include modern LLM design elements like Direct Preference Optimization (DPO) and Mixture-of-Experts (MoE) architectures. Liu et al. \cite{liu2025uncertainty} present a recent survey exploring the sources of uncertainty in LLMs and methods for uncertainty quantification and confidence calibration. Further, benchmarking efforts by researchers \cite{mucsanyi2024benchmarking} utilize conformal prediction to assess uncertainty, revealing insights into the relationship between model accuracy and uncertainty. Studies by Yona et al. \cite{yona2024can} and Xiong et al. \cite{xiong2024llmsexpressuncertaintyempirical} investigate the expression of uncertainty in LLMs, focusing on linguistic verbal uncertainty and numerical verbal confidence, respectively, under specific conditions.

\clearpage

\section{Comparison with Multi-Generation Based Uncertainty}
\label{appendix:semanticuncertainty}
We primarily compare two multi-generation based uncertainty estimation methods: (1) \emph{consistency}~\cite{kuhn2023semanticuncertaintylinguisticinvariances,manakul2023selfcheckgpt,yona2024can}, which measures the rate of contradiction between the anchor response and a set of sampled responses; and (2) \emph{majority vote}~\cite{farquhar2024detecting}, which estimates prediction confidence as the proportion of sampled responses that agree with the majority answer.

Since our evaluation is conducted under a multiple-choice setting, we focus only on the predicted label from each response, disregarding its full textual content.

\paragraph{Consistency}~Formally, given a question $Q$, a generated response $\hat{R}$, and its corresponding predicted answer $\hat{A}$, we sample $k$ additional responses $\{R_1, \ldots, R_k\}$ and extract their respective predictions $\{A_1, \ldots, A_k\}$. The model's uncertainty in $\hat{A}$ is then quantified as the fraction of sampled predictions that contradict $\hat{A}$:

\begin{equation}
    \text{unc}(\hat{A}) \equiv \frac{1}{k} \sum_{i=1}^{k} \mathbf{1}[\hat{A} \text{ contradicts } A_i]
    \label{eq:confidence_measure}
\end{equation}

To ensure a fair comparison, we use the same response $\hat{R}$ as those employed in LVU, NVU, and TPU, and set $k = 10$.

\paragraph{Majority Vote}~As an additional baseline, we adopt a majority vote strategy following~\cite{farquhar2024detecting}. In our multiple-choice setting, we omit the semantic clustering step and define the predicted label as the most frequent answer among the $k$ sampled responses. The proportion of responses that agree with this majority label is treated as the model's confidence.

Note that for models that fail to follow instructions (i.e., do not produce a valid prediction), we exclude the corresponding instances from evaluation.

\begin{figure}[h]
    \centering
    \includegraphics[width=\linewidth]{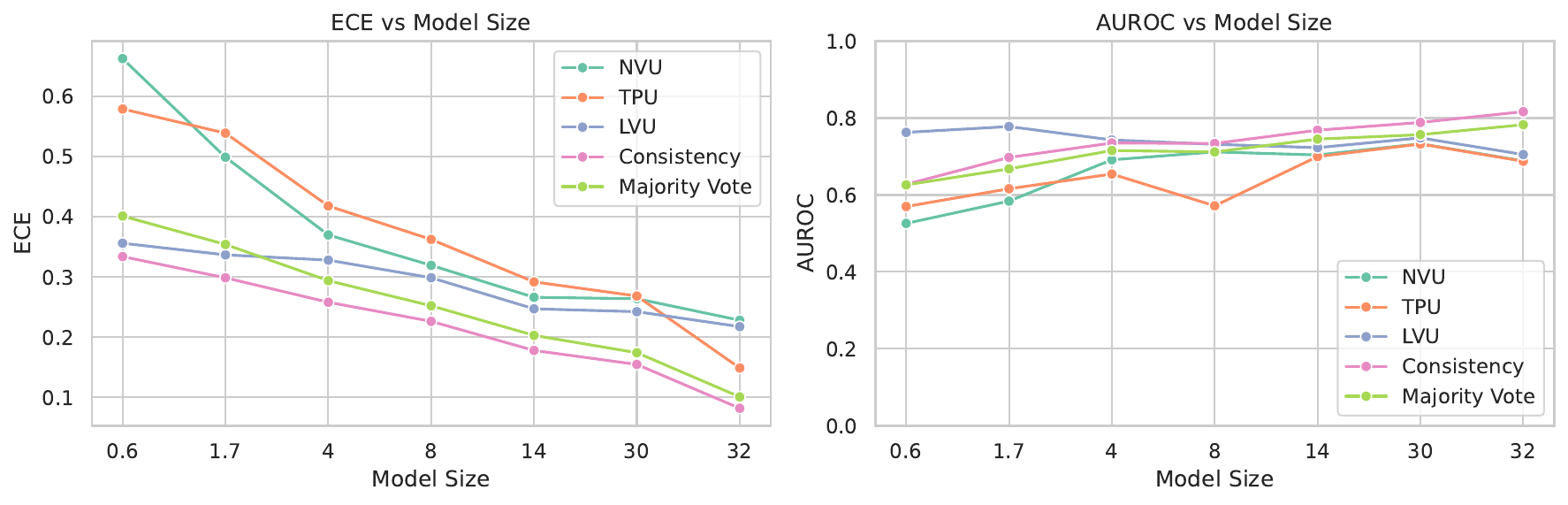}
    \caption{Comparison with multi-generation based uncertainty estimation.}
    \label{fig:Comparison with multi-generation based uncertainty estimation}
\end{figure}

We conduct experiments on the Qwen3 model series, ranging from 0.6B to 32B parameters, post trained models (with instruct fine tunning and DPO) and quantized variants (FP8). As shown in Figure~\ref{fig:Comparison with multi-generation based uncertainty estimation}, we compare post-trained Qwen3 series using multi-generation-based methods (Consistency and Majority Vote) and single-generation methods (LVU, NVU, and TPU). Overall, multi-generation approaches tend to yield better uncertainty estimation. In particular, Consistency consistently outperforms others in terms of calibration, achieving an ECE as low as 0.08 at the 32B model scale. However, the improvement in AUROC is relatively marginal.

Full evaluation results on multi-generation based uncertainty are provided in~\ref{Full evaluation results of semantic uncertainty}.

\clearpage
\subsection{Full evaluation results of multi-generation based uncertainty}
\label{Full evaluation results of semantic uncertainty}
\begin{table}[h]
    \caption{Multi-generation based uncertainty estimation evaluation for Qwen3 series.}
    \tiny
\begin{tabular}{l|c|c|ccccc|ccccc}
\toprule
                                 Model & Accuracy&Accuracy&  & & ECE & &  &   & & AUROC  & &  \\
                                 & &(Maj. Vote)& NVU & TPU & LVU & Consi. & Maj. Vote& NVU & TPU & LVU & Consi. & Maj. Vote \\
                                 \midrule
                                 Qwen3-0.6b&	0.248&	0.250&	0.663&	0.579&	0.356&	0.334&	0.401&	0.526&	0.570&	0.763&	0.628&	0.626\\
Qwen3-1.7b&	0.399&	0.418&	0.499&	0.539&	0.337&	0.299&	0.354&	0.584&	0.616&	0.778&	0.698&	0.668\\
Qwen3-4b&	0.529&	0.544&	0.370&	0.418&	0.328&	0.258&	0.294&	0.691&	0.654&	0.743&	0.735&	0.716\\
Qwen3-8b&	0.587&	0.602&	0.320&	0.363&	0.299&	0.226&	0.252&	0.712&	0.572&	0.732&	0.734&	0.712\\
Qwen3-14b&	0.632&	0.649&	0.266&	0.292&	0.247&	0.178&	0.203&	0.704&	0.700&	0.723&	0.768&	0.745\\
Qwen3-30b-a3b&	0.662&	0.686&	0.264&	0.269&	0.243&	0.155&	0.174&	0.733&	0.733&	0.748&	0.789&	0.757\\
Qwen3-32b&	0.697&	0.723&	0.228&	0.149&	0.218&	0.082&	0.101&	0.689&	0.688&	0.705&	0.817&	0.783\\
\midrule
Qwen3-0.6b-fp8	&0.250	&0.252	&0.635	&0.583	&0.369	&0.311	&0.398	&0.532	&0.570	&0.749	&0.631	&0.627\\
Qwen3-1.7b-fp8	&0.395	&0.416	&0.472	&0.540	&0.341	&0.277	&0.349	&0.611	&0.612	&0.770	&0.704	&0.671\\
Qwen3-4b-fp8	&0.530	&0.551	&0.368	&0.417	&0.329	&0.243	&0.284	&0.689	&0.652	&0.734	&0.736	&0.705\\
Qwen3-8b-fp8	&0.588	&0.613	&0.323	&0.361	&0.301	&0.208	&0.238	&0.708	&0.571	&0.730	&0.750	&0.714\\
Qwen3-14b-fp8	&0.630	&0.649	&0.269	&0.291	&0.245	&0.172	&0.199	&0.706	&0.699	&0.733	&0.768	&0.741\\
\bottomrule
    \end{tabular}
\end{table}

\section{Code and Resources}
Our code and resources are anonymously available at: \hyperlink{link}{https://anonymous.4open.science/r/LLM-Calibration-Study-203D/}

\clearpage

\section{Evaluation with Adaptive ECE}
\label{appendix:adaptiveece}
We also adopt Adaptive-ECE~\cite{nixon2019measuring} as an alternative metric to evaluate calibration performance. The results of Adaptive-ECE closely align with those of ECE, demonstrating the robustness of our calibration evaluation using ECE across different binning strategies in this work.

\begin{figure}[h]
    \centering
    \includegraphics[width=\linewidth]{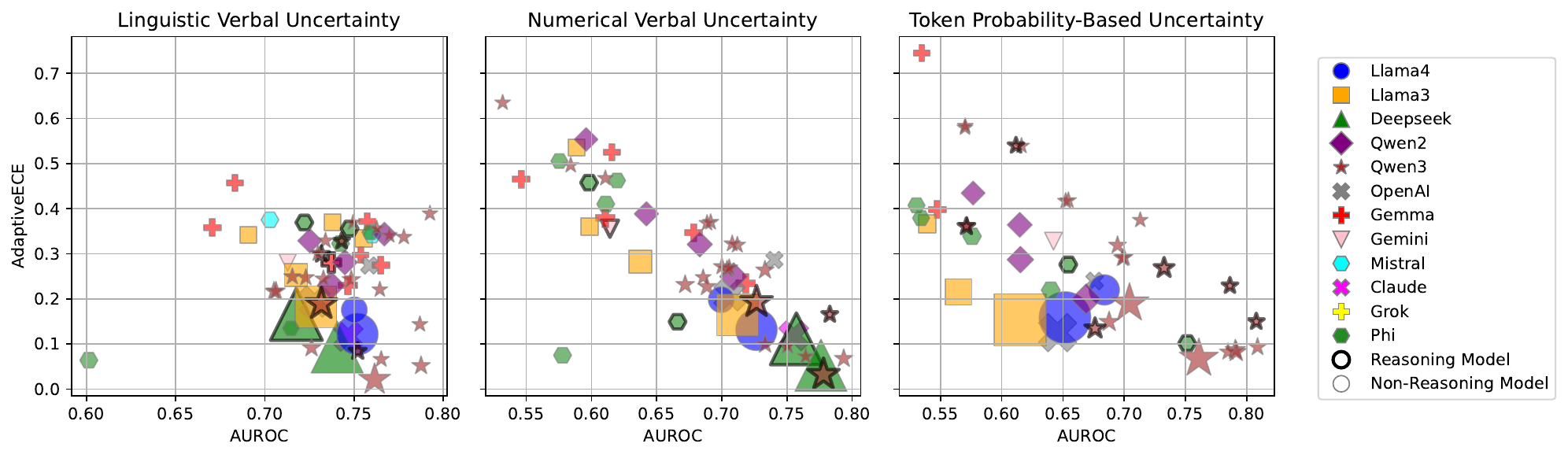}
    \caption{AdaptiveECE vs AUROC across all methods.}
    \label{fig:AdaptiveECE vs AUROC across all methods}
\end{figure}

\section{High Accuracy Does Not Guarantee Reliable Uncertainty Estimation}
\label{appendix:High Accuracy Does Not Guarantee Reliable Uncertainty Estimation}

\begin{figure}[h]
    \centering
    \includegraphics[width=0.7\linewidth]{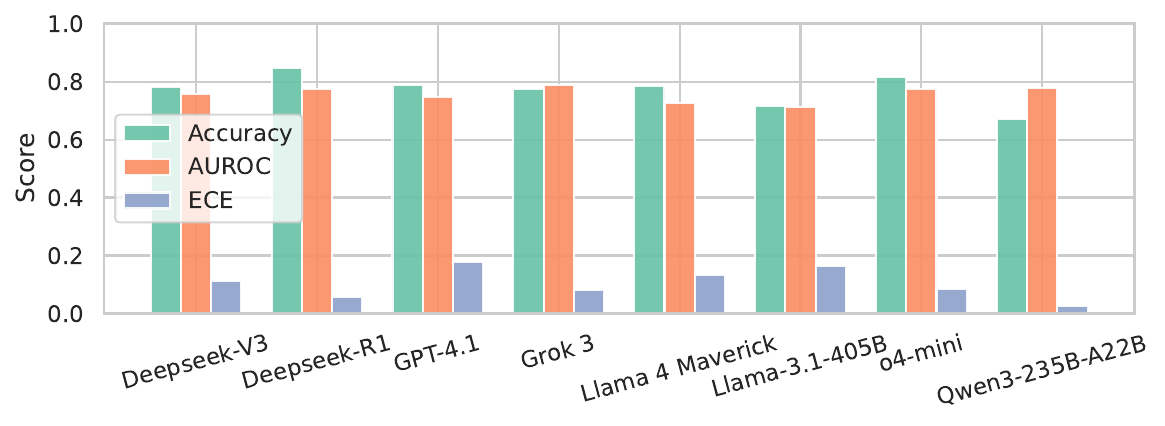}
    \caption{Comparison of top-tier models under NVU.}
    \label{fig:Comparison of top-tier models under NVU}
\end{figure}

\begin{figure}[h]
    \centering
    \includegraphics[width=0.7\linewidth]{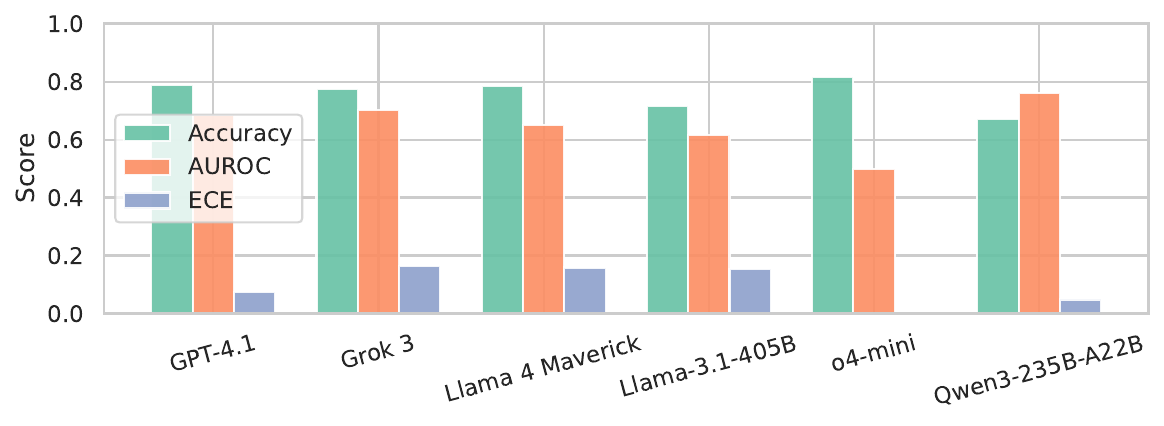}
    \caption{Comparison of top-tier models under TPU.}
    \label{fig:Comparison of top-tier models under TPU}
\end{figure}

As discussed in Section~\ref{main:High Accuracy Does Not Guarantee Reliable Uncertainty Estimation}, our experiments using LVU reveal that high accuracy does not necessarily imply reliable uncertainty estimation. We observe similar trends under NVU and TPU, as illustrated in Figures~\ref{fig:Comparison of top-tier models under NVU} and~\ref{fig:Comparison of top-tier models under TPU}. Notably, although \texttt{Qwen3-235B-A22B} exhibits the lowest accuracy among the evaluated models, it consistently achieves the highest AUROC and lowest ECE across all three uncertainty estimation methods. Note that the DeepSeek series does not provide valid token-level probabilities; hence, its AUROC and ECE values under TPU are omitted due to invalidity.

\clearpage

\section{Prompts and Regular Expressions}
\label{appendix:Prompts and Regular Expressions}

\subsection{Prompt Sharing for MMLU-PRO Response Sampling}
The following prompt is used to sample responses, incorporating NVU and LVU elicitation as well as a concise CoT, where \texttt{\{Letters\}} and \texttt{\{Question\}} are placeholders. A single response from this prompt is shared between and used by all uncertainty methods. 

\begin{tcolorbox}[colback=white, colframe=black, title=\textbf{Shared prompt for MMLU-Pro response sampling}, fonttitle=\bfseries, boxrule=0.8pt]

Read the question, analyze step by step, provide your answer and  your confidence in this answer. Note: The confidence indicates how  likely you think your answer is true.\\

If you are uncertain about your answer to the question, convey this  uncertainty linguistically by precisely hedging this answer. Use the following format to answer:\\

Explanation: [insert short step-by-step analysis here]\\
Answer: [ONLY the option LETTER, where the LETTER is one of \texttt{\{Letters\}}]\\
Confidence (0-100): [Your confidence level, please only include the numerical number in the range of 0-100]\%\\

Only give me the reply according to this format, don't give me any other words.\\
\\
\texttt{\{Question\}}
\end{tcolorbox}

\clearpage
\subsection{Prompts for Linguistic Uncertainty (LVU) Judge}
The following details the few-shot prompt used to instruct the LVU judge, where {QUESTION} and {Response} are placeholders. The response sampled from the shared prompt has been post-processed to remove numerical verbal confidence to avoid leading or biased LVU judgment. 

\begin{tcolorbox}[colback=white, colframe=black, title=\textbf{Prompt for the LVU judge}, fonttitle=\bfseries, boxrule=0.8pt]
You will be given a question and a proposed answer generated by a language model. Your task is to output an confidence score: a number in [0,1] representing how  certain the language model is about the truthfulness of its generated answer, judged solely based on the provided answer. You should pay special attention to the usage of any hedging modifiers in the original answer, used to convey uncertainty in the truthfulness of the answer. A score close to 1.0 indicates that the model is very certain about the truthfulness answer. And a score close to 0.0 means that the model is highly uncertain, and the answer likely contains some hedging expressions. If the proposed answer punts the question, the confidence score should be 0.0. \\

Here are some examples:\\

\textbf{Question}: criminal minds episode where jj becomes a profiler\\
\textbf{Proposed answer:} I think JJ officially becomes a profiler in Criminal Minds season 6, episode 2.\\
Confidence score: 0.8\\

\textbf{Question:} criminal minds episode where jj becomes a profiler \\
\textbf{Proposed answer:} JJ officially becomes a profiler in Criminal Minds season 6, in either the first or second episode.\\
Confidence score: 0.5\\

\textbf{Question:} criminal minds episode where jj becomes a profiler \\
\textbf{Proposed answer:} I'm not really sure about this, but I think the episode in which JJ officially becomes a profiler in Criminal Minds may be episode 2 in season 6.\\
Confidence score: 0.6\\

\textbf{Question:} criminal minds episode where jj becomes a profiler \\
\textbf{Proposed answer:} I don't know which episode you're referring to.\\
Confidence score: 0.0\\

Here are the question and proposed answer to be judged:\\
Question:\\
\texttt{\{Question\}}\\

Proposed Answer:\\
\texttt{\{Response\}}\\

Provide your response only in the following format:\\
\textbf{Confidence score:} [confidence score (0-1)].
\end{tcolorbox}

\clearpage
\subsection{Answer and Numerical Verbal Confidence Extraction Regular Expressions}
The following presents the regex collection used to extract the answer from the LLM responses. 

\begin{tcolorbox}[colback=white, colframe=black, title=\textbf{Multiple-choice answer extraction regex}, fonttitle=\bfseries, boxrule=0.8pt]
\begin{verbatim}
r"[Aa]nswer:?[\s]*[\n]*([A-J])",   
r"[Aa]nswer:[\s]*[\n]*\(?([A-J])\)?", 
r"[Aa]nswer:[\s]*[\n]*\[?([A-J])\]?",  
r"[Aa]nswer:[\s]*[\n]*([A-J])[,)]",         
r"[Aa]nswer:[\s]*[\n]*([A-J])\s*,?.*",
r"Answer:\n([A-J])\nConfidence",         
r"answer is\s*\[?\(?([A-J])\]?\)?",   
r"answer should be\s*\[?\(?([A-J])\]?\)?",   
r"best option is \(?([A-J])\)?",
r"best match is option \(?([A-J])\)?",
r"the closest is \(?([A-J])\)?",
r"Answer:\n*^([A-J])$",
r"^([A-J])$"
\end{verbatim}
\end{tcolorbox}

The following regex collection is used to extract and remove Numerical Verbal Confidence from the responses.
\begin{tcolorbox}[colback=white, colframe=black, title=\textbf{Numerical Verbal Confidence extraction and removal regex}, fonttitle=\bfseries, boxrule=0.8pt]
\begin{verbatim}
r"[Cc]onfidence\s*\(0-100\):\s*[\(]?[\[]?(\d+)[\)]?[\]]?%?", 
r"[Cc]onfidence[:]?\s*(\d+)%?",   
r"[Cc]onfidence [\(0-100\)]?:\s*\[(\d+)%?\]"
r"[Cc]onfidence [Ll]evel\s*\(0-100\):\s*(\d+)%?", 
r"[Cc]onfidence [Ll]evel[:]?\s*(\d+)%?",      
r"[Cc]onfidence [Ll]evel[\(0-100\)]?:\s*\[(\d+)%?\]",
r"[Cc]onfidence \(100\):\s*\w*,\s*(\d+)%?",
r"[Cc]onfidence\s*\(\d+\)\s*:\s*(\d+)%?",
r"[Cc]onfidence\s*[\(]?(\d+)[\)]?%?"
\end{verbatim}
\end{tcolorbox}

Any output falling outside the predefined regex patterns is treated as non-response, even if the underlying content may be partially or fully accurate. As a result, correct answers and successful confidence reporting are contingent upon both factual correctness and adherence to basic instruction-following behavior. 

Our observations suggest that answer and confidence extraction failures occur due to poor instruction following, and the models often exhibit the following behavior.

\begin{itemize}
    \item Implicit answers: the content of the response is correct, and yet the answer is not explicitly stated.
    \item Maximum token length exceeded: especially evident in reasoning models, the models have not finished outputting their chain of thought before hitting the maximum token length. i.e. the answer and/or confidence have not yet been provided. 
    \item Non-response error: due to model internal issues, null responses are produced arbitrarily. 
    \item General regex mismatches: the answer output does not adhere to the allowed regex formats. 
\end{itemize}

For subsequent analyses, all successful confidence extractions will be rescaled between 0 - 1. All failed extractions will be dropped. On average, ~17\% of the responses are dropped due to the answer, numerical confidence and/or linguistic confidence extraction failure. 

\clearpage

\section{Efficient Confidence Evaluation: Stratified Sampling on MMLU-Pro}
Since instruction following can be challenging for less-capable models such as Qwen3-0.6B, we remove instances where the model fails to adhere to the instructions or produces nonsensical responses. For example, Qwen3-0.6B has up to 5{,}000 such invalid responses excluded from evaluation. To ensure that our confidence estimation remains reliable despite the reduced number of valid responses, we conduct additional experiments demonstrating that the evaluation performance remains consistent even with fewer samples.

Although MMLU-Pro comprises a substantial total of 12,032 questions spanning a diverse range of academic disciplines, we are interested in the possibility of evaluating confidence on a subset to reduce computational overhead without significantly compromising the reliability of the results. 

\begin{figure}[h]
    \centering
    \includegraphics[width=\linewidth]{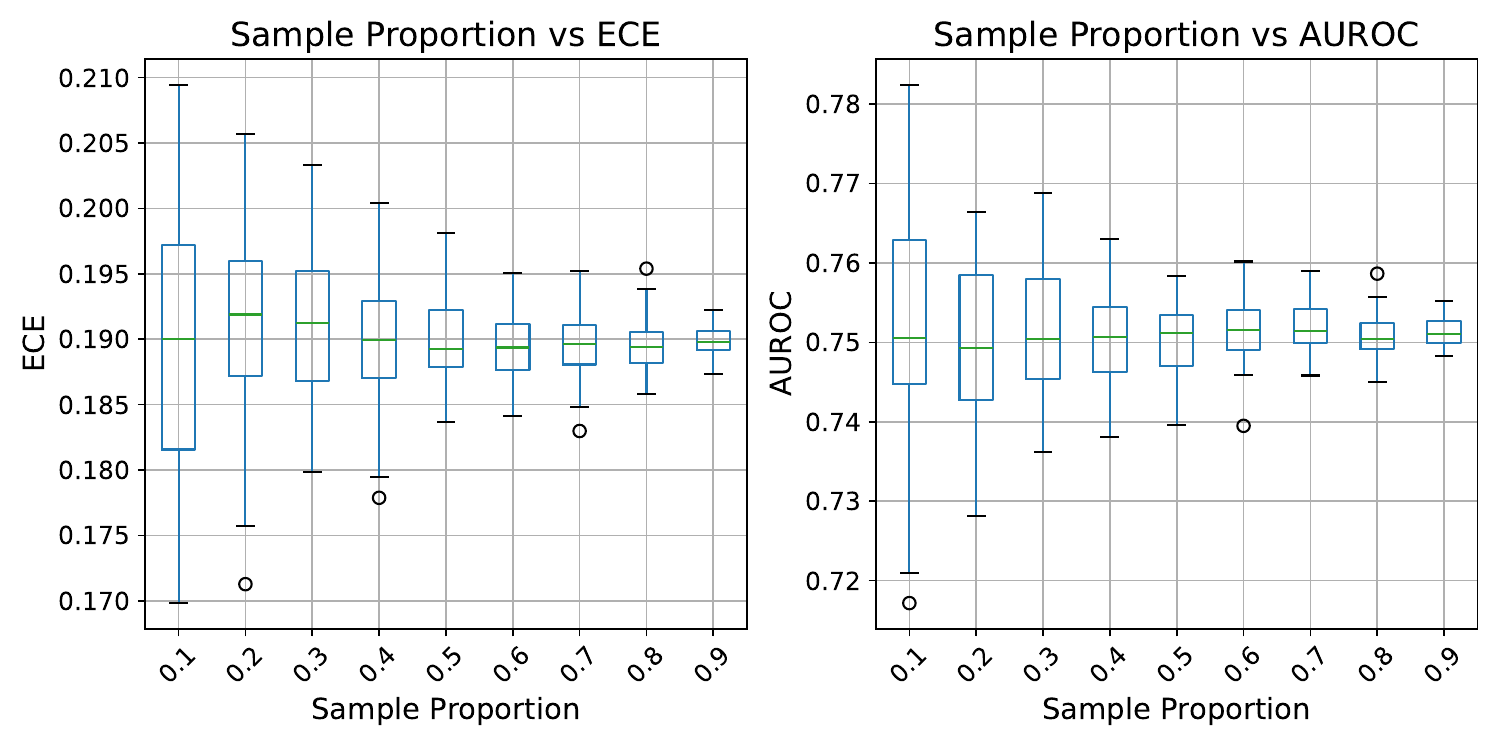}
    \caption{\textbf{GPT-4.1-Mini's spread of confidence distributions from different stratified subsets sizes.} Each box represents 50 stratified samples. The AUROC and ECE distributions start to stabilize and trend towards those of the full set when the sample proportion exceeds 50\%. }
    \label{fig:mmlu_pro_subsets}
\end{figure}

To investigate this, we employ GPT-4.1-Mini to generate responses for the entire MMLU-Pro dataset and construct stratified samples at proportions of from 10\% to 90\%. Each sample is generated through stratified sampling, ensuring an equal number of questions from each field of study. For each sampling proportion, we repeat the procedure 50 times to account for sampling variability and to obtain robust performance estimates. The resulting distributions of ECE and AUROC are presented in Figure~\ref{fig:mmlu_pro_subsets}. The results demonstrate that stratified samples comprising at least 50\% of the dataset produce ECE and AUROC distributions that are closely aligned with those of the full dataset, thereby validating that the use of stratified subsets are applicable for efficient calibration analysis. 

\clearpage

\section{LLM as a Judge for LVU}
\label{appendix:llmasajudge}
We employ LLaMa-4-Maverick-17B-128E~\cite{meta2025llama4} as the judge for LVU in all evaluations. We replicate the experiment carried out by \cite{belem2024perceptions} to explore the judge's mapping between various hedging words and their perceived numerical confidence (note that confidence and uncertainty are complementary, i.e., $ \text{uncertainty} = 1 - \text{confidence}$). 
The experiment involves evaluating a collection of sentences, each of which consists of a single hedging word, and outputting a corresponding probability of truth ascertained by the model. There are 20 non-verifiable statements on daily life events, where non-verifiable statements are statements for which there is insufficient contextual information for external observers to confidently believe whether they are true or false. This is designed to linguistically assess the model's sensitivity to hedging words rather than its knowledge. 

\subsection{Sentence Construction}
Each sentence has the following structure:
\begin{tcolorbox}[colback=white, colframe=black, title=\textbf{Sentence structure}, fonttitle=\bfseries, boxrule=0.8pt]
\texttt{[NAME]} believes it is \texttt{[HEDGING WORD]} that \texttt{[STATEMENT]}
\end{tcolorbox}
where the name, hedging word, and statement are sourced from the following. The judge then evaluates all the possible combinations of names, hedging words, and statements.

\begin{tcolorbox}[colback=white, colframe=black, title=\textbf{Possible options for [NAME]}, fonttitle=\bfseries, boxrule=0.8pt]
"Brendan", "Bruce", "David", "Gary", "Isaac", "Jeffery", 
"Joey", "Johnnie", "Kenny", "Lance", "Marco", "Mike", "Nathan", 
"Nick", "Raul", "Amanda", "Bonnie", "Camille", "Catherine", 
"Cheri", "Ethel", "Gabriela", "Jacquelyn", "Jessica",
"Laura", "Olga", "Roxanne", "Silvia", "Tara", "Violet"
\end{tcolorbox}

\begin{tcolorbox}[colback=white, colframe=black, title=\textbf{Possible options for [HEDGING WORD]}, fonttitle=\bfseries, boxrule=0.8pt]
"almost certain", "highly likely", "very likely", "probable", 
"somewhat likely", "possible", "uncertain", "somewhat unlikely", 
"unlikely", "not likely", "doubtful", "very unlikely", "highly unlikely"
\end{tcolorbox}

\begin{tcolorbox}[colback=white, colframe=black, title=\textbf{Possible options for [STATEMENT]}, fonttitle=\bfseries, boxrule=0.8pt]
"they will buy a new watch this Thanksgiving weekend.",\\
"they will be offered a promotion this fall.",\\
"the company will have another round of layoffs by mid-July.",\\
"there will be vegetarian options at the barbecue.",\\
"they will visit New York over winter break.",\\
"the restaurant near their apartment accepts reservations.",\\
"the new museum is offering complimentary admission.",\\
"there is a yoga studio within 2 miles of their workplace.",\\
"there are more than eighty students in the auditorium right now.",\\
"the temperature in the office is at least 72 degrees Fahrenheit.",\\
"their boss owns a blue car.",\\
"their friend has a leather jacket.",\\
"their cousin has a vegetable garden.",\\
"their classmate owns a guitar.",\\
"their boss has a stereo amplifier.",\\
"their cousin prefers spinach over broccoli.",\\
"their boss prefers coffee over tea.",\\
"their friend prefers running over cycling.",\\
"their neighbor prefers the beach over the mountains.",\\
"their coworker prefers reading books over watching movies."
\end{tcolorbox}

\subsection{Prompt Construction and Uncertainty Value Extraction}
The following prompt is used to elicit the judge to output its perceived uncertainty with the presence of a hedging word in the sentence. 
\begin{tcolorbox}[colback=white, colframe=black, title=\textbf{Prompt to elicit LLMs' perceived confidence}, fonttitle=\bfseries, boxrule=0.8pt]
\texttt{[SENTENCE]} \\

From their perspective, considering the use of hedging words, what is the probability (from 0 to 100) that the statement is true? Concisely provide your response only in the following format:\\

Probability: [0-100]%
\end{tcolorbox}

The probability is then extracted using the following regular expression:
\begin{tcolorbox}[colback=white, colframe=black, title=\textbf{Regex to extract perceived confidence}, fonttitle=\bfseries, boxrule=0.8pt]
    \begin{verbatim}
    r"\b(?:Probability|Prob(?:\.|ability)?|P)\s*[:=]?\s*(\d+)%"
    \end{verbatim}
\end{tcolorbox}

\subsection{Results}
The prompt and regex used above successfully capture 99.85\% of the sentences. The high success rate represents the great instruction following ability of the judge.

\begin{figure}[h]
    \centering
    \includegraphics[width=0.9\linewidth]{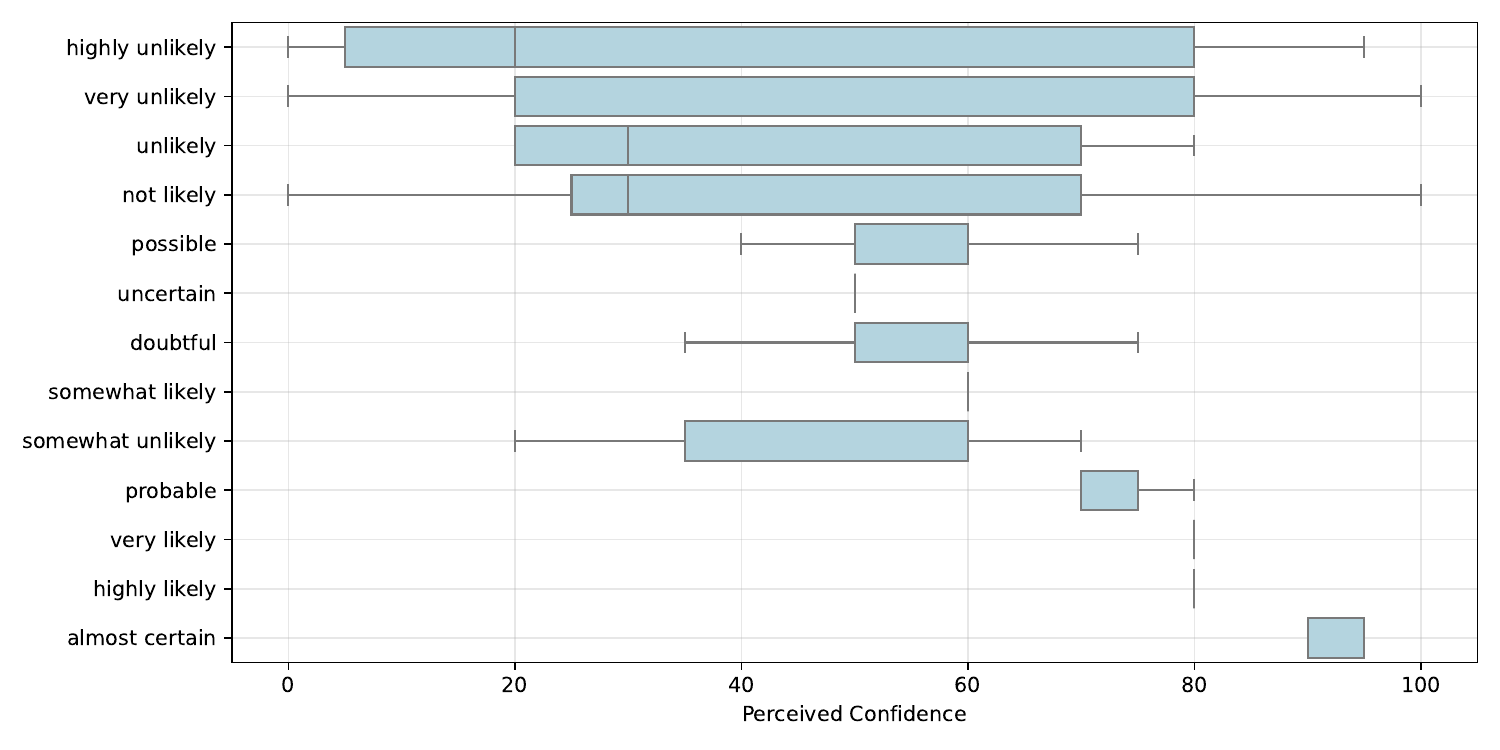}
    \caption{LLaMa-4-Maverick-17B-128E’s mapping of hedging words to LVU demonstrates that the model can distinguish between different relative levels of uncertainty. Specifically, perceived confidence increases consistently with the apparent confidence expressed in a sentence.}
    \label{fig:hedging_words}
\end{figure}

As shown in Figure \ref{fig:hedging_words}, the distribution patterns of linguistic expressions reveal a meaningful trend in the model’s responses, wherein stronger verbal hedges such as “almost certain” consistently receive higher confidence scores compared to weaker expressions like “probably not” or “unlikely.” This indicates that the model demonstrates a robust understanding of the semantic gradation embedded in hedging language. In particular, more decisive hedges—such as “almost certain” and “highly likely”—are interpreted with high confidence, reflecting a clear semantic alignment with human intuitions regarding probability and certainty. Conversely, more negative hedges such as “highly unlikely” and “probably not” exhibit lower values in their associated confidence scores. Such characteristics make LLaMa-4-Maverick-17B-128E a reliable linguistic judge for uncertainty evaluations. 

\clearpage

\section{Full Evaluation Results}
\label{appendix:evaluation results}
\input{eval_results}

%% file: eval_results.tex
\begin{table}[!ht]
    \centering
        \caption{\textbf{Full Evaluation Results (Part 1)}~Model naming format follows HuggingFace.}
    \small
    \begin{tabular}{lcccccccc}
    \toprule
        \textbf{Model} & \textbf{Accuracy} &
        \multicolumn{3}{c}{\textbf{ECE}} &
        \multicolumn{3}{c}{\textbf{AUROC}} \\
        \cmidrule(lr){3-5} \cmidrule(lr){6-8}
        & &
        \textbf{NVU} &
        \textbf{TPU} &
        \textbf{LVU} &
        \textbf{NVU} &
        \textbf{TPU} &
        \textbf{LVU} \\

        \midrule
        \textbf{Claude-3-5-haiku-20241022} & 0.603 & 0.297 & / & 0.272 & 0.676 & / & 0.718 \\ 
        \textbf{Claude-3-7-sonnet-20250219} & 0.782 & 0.119 & / & 0.118 & 0.755 & / & 0.747 \\ 
        \textbf{Claude-3-haiku-20240307} & 0.421 & 0.498 & / & 0.397 & 0.525 & / & 0.649 \\ 
        \midrule
        \textbf{Deepseek-chat} & 0.784 & 0.112 & / & 0.166 & 0.757 & / & 0.718 \\ 
        \textbf{Deepseek-reasoner} & 0.846 & 0.055 & / & 0.094 & 0.776 & / & 0.741 \\ 
        \midrule
        \textbf{Gemini-1.5-flash} & 0.650 & 0.273 & 0.275 & 0.238 & 0.687 & 0.683 & 0.724 \\ 
        \textbf{Gemini-1.5-flash-8b} & 0.542 & 0.358 & 0.328 & 0.281 & 0.614 & 0.642 & 0.713 \\ 
        \textbf{Gemini-1.5-pro} & 0.741 & 0.207 & 0.189 & 0.188 & 0.674 & 0.661 & 0.705 \\ 
        \textbf{Gemini-2.0-flash} & 0.785 & 0.177 & 0.091 & 0.162 & 0.721 & 0.707 & 0.744 \\ 
        \textbf{Gemini-2.0-flash-lite} & 0.723 & 0.207 & 0.138 & 0.182 & 0.720 & 0.724 & 0.756 \\ 
        \midrule
        \textbf{Gemma-2-27b-it} & 0.533 & 0.380 & 0.351 & 0.281 & 0.611 & 0.524 & 0.737 \\ 
        \textbf{Gemma-2-9b-it} & 0.470 & 0.465 & 0.398 & 0.358 & 0.546 & 0.547 & 0.671 \\ 
        \textbf{Gemma-2b-it} & 0.148 & 0.767 & 0.745 & 0.457 & 0.504 & 0.535 & 0.683 \\ 
        \textbf{Gemma-3-12b-it} & 0.607 & 0.347 & / & 0.275 & 0.678 & / & 0.765 \\ 
        \textbf{Gemma-3-1b-it} & 0.175 & 0.782 & 0.694 & 0.298 & 0.513 & 0.514 & 0.754 \\ 
        \textbf{Gemma-3-27b-it} & 0.667 & 0.234 & / & 0.230 & 0.718 & / & 0.747 \\ 
        \textbf{Gemma-3-4b-it} & 0.430 & 0.525 & / & 0.372 & 0.616 & / & 0.757 \\ 
        \midrule
        \textbf{Gpt-4.1} & 0.791 & 0.176 & 0.074 & 0.166 & 0.748 & 0.684 & 0.748 \\ 
        \textbf{Gpt-4.1-mini} & 0.761 & 0.168 & 0.130 & 0.184 & 0.765 & 0.727 & 0.744 \\ 
        \textbf{Gpt-4.1-nano} & 0.635 & 0.243 & 0.214 & 0.252 & 0.712 & 0.698 & 0.781 \\ 
        \textbf{Gpt-4o-2024-08-06} & 0.726 & 0.208 & 0.125 & 0.194 & 0.706 & 0.646 & 0.728 \\ 
        \textbf{Gpt-4o-mini-2024-07-18} & 0.601 & 0.286 & 0.240 & 0.274 & 0.740 & 0.676 & 0.759 \\
        \midrule
        \textbf{Grok-2-1212} & 0.726 & 0.215 & 0.200 & 0.198 & 0.670 & 0.613 & 0.712 \\ 
        \textbf{Grok-3-beta} & 0.774 & 0.081 & 0.164 & 0.097 & 0.790 & 0.703 & 0.774 \\ 
        \textbf{Grok-3-mini-beta} & 0.829 & 0.102 & 0.138 & 0.101 & 0.791 & 0.676 & 0.783 \\ 
        \midrule
        \textbf{Llama-3.1-nemotron-nano-4b-v1.1} & 0.502 & 0.332 & 0.350 & 0.285 & 0.721 & 0.662 & 0.762 \\ 
        \textbf{Llama-3.1-nemotron-nano-8b-v1} & 0.401 & 0.349 & 0.298 & 0.302 & 0.637 & 0.628 & 0.707 \\ 
        \textbf{Llama-3.2-1b-instruct} & 0.166 & 0.510 & 0.573 & 0.342 & 0.525 & 0.463 & 0.691 \\ 
        \textbf{Llama-3.2-3b-instruct} & 0.317 & 0.536 & 0.476 & 0.370 & 0.589 & 0.521 & 0.738 \\ 
        \textbf{Llama-3.1-8b-instruct} & 0.449 & 0.361 & 0.366 & 0.334 & 0.599 & 0.539 & 0.755 \\ 
        \textbf{Llama-3.1-70b-instruct} & 0.610 & 0.282 & 0.215 & 0.253 & 0.638 & 0.565 & 0.718 \\ 
        \textbf{Llama-3.1-405b-instruct} & 0.722 & 0.165 & 0.153 & 0.183 & 0.712 & 0.615 & 0.729 \\ 
        
        \textbf{Llama-4-maverick-17b-128e-instruct-fp8} & 0.787 & 0.131 & 0.158 & 0.122 & 0.727 & 0.652 & 0.752 \\ 
        \textbf{Llama-4-scout-17b-16e-instruct} & 0.740 & 0.198 & 0.220 & 0.176 & 0.700 & 0.684 & 0.750 \\ 
        \midrule
        \textbf{Mistral-7b-instruct-v0.1} & 0.263 & 0.677 & 0.606 & 0.341 & 0.495 & 0.486 & 0.760 \\ 
        \textbf{Mistral-nemo-instruct-2407} & 0.405 & 0.580 & 0.446 & 0.375 & 0.522 & 0.510 & 0.703 \\ 
        \textbf{Mistral-small-24b-instruct-2501} & 0.598 & 0.309 & 0.275 & 0.283 & 0.651 & 0.544 & 0.696 \\ 
        \textbf{Mistral-small-3.1-24b-base-2503} & 0.545 & 0.354 & 0.308 & 0.244 & 0.552 & 0.471 & 0.624 \\ 
        \midrule
                        \textbf{O1-mini-2024-09-12} & 0.713 & 0.218 & / & 0.210 & 0.776 & / & 0.777 \\ 
        \textbf{O3-mini-2025-01-31} & 0.797 & 0.161 & / & 0.152 & 0.757 & / & 0.752 \\ 
        \textbf{O4-mini-2025-04-16} & 0.832 & 0.083 & / & 0.106 & 0.776 & / & 0.738 \\ 
        \midrule
        \textbf{Phi-3-medium-128k-instruct} & 0.504 & 0.410 & 0.339 & 0.323 & 0.611 & 0.576 & 0.742 \\ 
        \textbf{Phi-3-mini-128k-instruct} & 0.399 & 0.506 & 0.407 & 0.356 & 0.575 & 0.530 & 0.748 \\ 
        \textbf{Phi-3.5-mini-instruct} & 0.442 & 0.462 & 0.379 & 0.370 & 0.620 & 0.534 & 0.722 \\ 
        \textbf{Phi-3.5-moe-instruct} & 0.518 & 0.383 & 0.307 & 0.313 & 0.649 & 0.581 & 0.746 \\ 
        \textbf{Phi-4-mini-instruct} & 0.460 & 0.458 & 0.276 & 0.348 & 0.598 & 0.654 & 0.759 \\ 
        \textbf{Phi-4-mini-reasoning} & 0.801 & 0.149 & 0.094 & 0.135 & 0.666 & 0.751 & 0.715 \\ 
        \textbf{Phi-4-reasoning} & 0.915 & 0.075 & 0.219 & 0.064 & 0.578 & 0.640 & 0.601 \\ 
              \bottomrule
    \end{tabular}
\end{table}

\begin{table}[!ht]
    \caption{\textbf{Full Evaluation Results (Part 2)}~Model naming format follows HuggingFace.}
    \centering
    \small
    \begin{tabular}{lcccccccc}
    \toprule
        \textbf{Model} & \textbf{Accuracy} &
        \multicolumn{3}{c}{\textbf{ECE}} &
        \multicolumn{3}{c}{\textbf{AUROC}} \\
        \cmidrule(lr){3-5} \cmidrule(lr){6-8}
        & &
        \textbf{NVU} &
        \textbf{TPU} &
        \textbf{LVU} &
        \textbf{NVU} &
        \textbf{TPU} &
        \textbf{LVU} \\
        \midrule
        \textbf{Qwen2.5-14b-instruct} & 0.571 & 0.321 & 0.287 & 0.281 & 0.683 & 0.615 & 0.745 \\ 
        \textbf{Qwen2.5-32b-instruct} & 0.663 & 0.251 & 0.200 & 0.230 & 0.709 & 0.669 & 0.737 \\ 
        \textbf{Qwen2.5-3b-instruct} & 0.396 & 0.554 & 0.435 & 0.344 & 0.596 & 0.577 & 0.767 \\ 
        \textbf{Qwen2.5-7b-instruct} & 0.508 & 0.389 & 0.364 & 0.329 & 0.642 & 0.615 & 0.725 \\ 
        \textbf{Qwen3-0.6b} & 0.248 & 0.663 & 0.579 & 0.356 & 0.526 & 0.570 & 0.763 \\ 
        \textbf{Qwen3-0.6b-base} & 0.275 & 0.711 & 0.546 & 0.321 & 0.501 & 0.575 & 0.784 \\ 
        \textbf{Qwen3-0.6b-fp8} & 0.250 & 0.635 & 0.583 & 0.369 & 0.532 & 0.570 & 0.749 \\ 
        \textbf{Qwen3-0.6b-think} & 0.415 & 0.556 & 0.375 & 0.389 & 0.519 & 0.713 & 0.793 \\ 
        \textbf{Qwen3-1.7b} & 0.399 & 0.499 & 0.539 & 0.337 & 0.584 & 0.616 & 0.778 \\ 
        \textbf{Qwen3-1.7b-base} & 0.374 & 0.539 & 0.435 & 0.334 & 0.538 & 0.560 & 0.776 \\ 
        \textbf{Qwen3-1.7b-fp8} & 0.395 & 0.472 & 0.540 & 0.341 & 0.611 & 0.612 & 0.770 \\ 
        \textbf{Qwen3-1.7b-think} & 0.695 & 0.249 & 0.224 & 0.221 & 0.686 & 0.786 & 0.764 \\ 
        \textbf{Qwen3-14b} & 0.632 & 0.266 & 0.292 & 0.247 & 0.704 & 0.700 & 0.723 \\ 
        \textbf{Qwen3-14b-awq} & 0.615 & 0.273 & 0.319 & 0.249 & 0.700 & 0.695 & 0.715 \\ 
        \textbf{Qwen3-14b-base} & 0.593 & 0.335 & 0.270 & 0.284 & 0.624 & 0.595 & 0.706 \\ 
        \textbf{Qwen3-14b-fp8} & 0.630 & 0.269 & 0.291 & 0.245 & 0.706 & 0.699 & 0.733 \\ 
        \textbf{Qwen3-14b-think} & 0.829 & 0.096 & 0.090 & 0.084 & 0.750 & 0.809 & 0.752 \\ 
        \textbf{Qwen3-235b-a22b-fp8-tput} & 0.728 & 0.193 & 0.190 & 0.188 & 0.727 & 0.704 & 0.732 \\ 
        \textbf{Qwen3-235b-a22b-fp8-tput-think} & 0.880 & 0.025 & 0.048 & 0.015 & 0.778 & 0.761 & 0.762 \\ 
        \textbf{Qwen3-30b-a3b} & 0.662 & 0.264 & 0.269 & 0.243 & 0.733 & 0.733 & 0.748 \\ 
        \textbf{Qwen3-30b-a3b-base} & 0.604 & 0.324 & 0.253 & 0.269 & 0.637 & 0.558 & 0.724 \\ 
        \textbf{Qwen3-30b-a3b-think} & 0.861 & 0.068 & 0.080 & 0.052 & 0.794 & 0.791 & 0.788 \\ 
        \textbf{Qwen3-32b} & 0.697 & 0.228 & 0.149 & 0.218 & 0.689 & 0.688 & 0.705 \\ 
        \textbf{Qwen3-32b-awq} & 0.711 & 0.232 & 0.135 & 0.217 & 0.672 & 0.676 & 0.706 \\ 
        \textbf{Qwen3-32b-think} & 0.841 & 0.099 & 0.042 & 0.092 & 0.733 & 0.791 & 0.726 \\ 
        \textbf{Qwen3-4b} & 0.529 & 0.370 & 0.418 & 0.328 & 0.691 & 0.654 & 0.743 \\ 
        \textbf{Qwen3-4b-base} & 0.526 & 0.403 & 0.324 & 0.329 & 0.643 & 0.573 & 0.735 \\ 
        \textbf{Qwen3-4b-fp8} & 0.530 & 0.368 & 0.417 & 0.329 & 0.689 & 0.652 & 0.734 \\ 
        \textbf{Qwen3-4b-think} & 0.766 & 0.165 & 0.142 & 0.144 & 0.783 & 0.808 & 0.787 \\ 
        \textbf{Qwen3-8b} & 0.587 & 0.320 & 0.363 & 0.299 & 0.712 & 0.572 & 0.732 \\ 
        \textbf{Qwen3-8b-base} & 0.556 & 0.359 & 0.310 & 0.307 & 0.660 & 0.565 & 0.728 \\ 
        \textbf{Qwen3-8b-fp8} & 0.588 & 0.323 & 0.361 & 0.301 & 0.708 & 0.571 & 0.730 \\ 
        \textbf{Qwen3-8b-think} & 0.840 & 0.073 & 0.072 & 0.068 & 0.764 & 0.785 & 0.765 \\ 
        \bottomrule
    \end{tabular}
\end{table}

%% file: neurips_2025.bbl
\begin{thebibliography}{54}
\providecommand{\natexlab}[1]{#1}
\providecommand{\url}[1]{\texttt{#1}}
\expandafter\ifx\csname urlstyle\endcsname\relax
  \providecommand{\doi}[1]{doi: #1}\else
  \providecommand{\doi}{doi: \begingroup \urlstyle{rm}\Url}\fi

\bibitem[Thirunavukarasu et~al.(2023)Thirunavukarasu, Ting, Elangovan, Gutierrez, Tan, and Ting]{thirunavukarasu2023large}
Arun~James Thirunavukarasu, Darren Shu~Jeng Ting, Kabilan Elangovan, Laura Gutierrez, Ting~Fang Tan, and Daniel Shu~Wei Ting.
\newblock Large language models in medicine.
\newblock \emph{Nature medicine}, 29\penalty0 (8):\penalty0 1930--1940, 2023.

\bibitem[Van~Veen et~al.(2024)Van~Veen, Van~Uden, Blankemeier, Delbrouck, Aali, Bluethgen, Pareek, Polacin, Reis, Seehofnerov{\'a}, et~al.]{van2024adapted}
Dave Van~Veen, Cara Van~Uden, Louis Blankemeier, Jean-Benoit Delbrouck, Asad Aali, Christian Bluethgen, Anuj Pareek, Malgorzata Polacin, Eduardo~Pontes Reis, Anna Seehofnerov{\'a}, et~al.
\newblock Adapted large language models can outperform medical experts in clinical text summarization.
\newblock \emph{Nature medicine}, 30\penalty0 (4):\penalty0 1134--1142, 2024.

\bibitem[Abacha et~al.(2024)Abacha, Yim, Fu, Sun, Yetisgen, Xia, and Lin]{abacha2024medec}
Asma~Ben Abacha, Wen-wai Yim, Yujuan Fu, Zhaoyi Sun, Meliha Yetisgen, Fei Xia, and Thomas Lin.
\newblock Medec: A benchmark for medical error detection and correction in clinical notes.
\newblock \emph{arXiv preprint arXiv:2412.19260}, 2024.

\bibitem[Hadi et~al.(2023)Hadi, Qureshi, Shah, Irfan, Zafar, Shaikh, Akhtar, Wu, Mirjalili, et~al.]{hadi2023large}
Muhammad~Usman Hadi, Rizwan Qureshi, Abbas Shah, Muhammad Irfan, Anas Zafar, Muhammad~Bilal Shaikh, Naveed Akhtar, Jia Wu, Seyedali Mirjalili, et~al.
\newblock Large language models: a comprehensive survey of its applications, challenges, limitations, and future prospects.
\newblock \emph{Authorea Preprints}, 1:\penalty0 1--26, 2023.

\bibitem[Lai et~al.(2024)Lai, Gan, Wu, Qi, and Yu]{lai2024large}
Jinqi Lai, Wensheng Gan, Jiayang Wu, Zhenlian Qi, and Philip~S Yu.
\newblock Large language models in law: A survey.
\newblock \emph{AI Open}, 2024.

\bibitem[Passi and Vorvoreanu(2022)]{passi2022overreliance}
Samir Passi and Mihaela Vorvoreanu.
\newblock Overreliance on ai literature review.
\newblock \emph{Microsoft Research}, 339:\penalty0 340, 2022.

\bibitem[{ISACA}(2024)]{isaca2024euai}
{ISACA}.
\newblock Understanding the eu ai act: Requirements and next steps.
\newblock \url{https://www.isaca.org/resources/white-papers/2024/understanding-the-eu-ai-act}, October 2024.
\newblock Accessed: 2025-05-12.

\bibitem[{BBC News}(2023)]{bbc2023debtceiling}
{BBC News}.
\newblock Us debt ceiling deal clears major hurdle in congress, 2023.
\newblock URL \url{https://www.bbc.com/news/world-us-canada-65735769}.
\newblock Accessed: 2025-05-10.

\bibitem[Duan et~al.(2023)Duan, Cheng, Wang, Zavalny, Wang, Xu, Kailkhura, and Xu]{duan2023shifting}
Jinhao Duan, Hao Cheng, Shiqi Wang, Alex Zavalny, Chenan Wang, Renjing Xu, Bhavya Kailkhura, and Kaidi Xu.
\newblock Shifting attention to relevance: Towards the uncertainty estimation of large language models.
\newblock 2023.

\bibitem[Lamb et~al.()Lamb, Ivanova, Torr, and Rudner]{lambsemantic}
Tom~A Lamb, Desi~R Ivanova, Philip Torr, and Tim~GJ Rudner.
\newblock Semantic calibration of llms through the lens of temperature scaling.
\newblock In \emph{ICLR Workshop: Quantify Uncertainty and Hallucination in Foundation Models: The Next Frontier in Reliable AI}.

\bibitem[Kossen et~al.(2024)Kossen, Han, Razzak, Schut, Malik, and Gal]{kossen2024semantic}
Jannik Kossen, Jiatong Han, Muhammed Razzak, Lisa Schut, Shreshth Malik, and Yarin Gal.
\newblock Semantic entropy probes: Robust and cheap hallucination detection in llms.
\newblock \emph{arXiv preprint arXiv:2406.15927}, 2024.

\bibitem[Ji et~al.(2025)Ji, Yu, Koishekenov, Bang, Hartshorn, Schelten, Zhang, Fung, and Cancedda]{ji2025calibrating}
Ziwei Ji, Lei Yu, Yeskendir Koishekenov, Yejin Bang, Anthony Hartshorn, Alan Schelten, Cheng Zhang, Pascale Fung, and Nicola Cancedda.
\newblock Calibrating verbal uncertainty as a linear feature to reduce hallucinations.
\newblock \emph{arXiv preprint arXiv:2503.14477}, 2025.

\bibitem[Zhu et~al.(2025)Zhu, Tao, Dong, and Xu]{zhu2025mitigating}
Younan Zhu, Linwei Tao, Minjing Dong, and Chang Xu.
\newblock Mitigating object hallucinations in large vision-language models via attention calibration.
\newblock \emph{arXiv preprint arXiv:2502.01969}, 2025.

\bibitem[Kuhn et~al.(2023)Kuhn, Gal, and Farquhar]{kuhn2023semanticuncertaintylinguisticinvariances}
Lorenz Kuhn, Yarin Gal, and Sebastian Farquhar.
\newblock Semantic uncertainty: Linguistic invariances for uncertainty estimation in natural language generation, 2023.
\newblock URL \url{https://arxiv.org/abs/2302.09664}.

\bibitem[Manakul et~al.(2023)Manakul, Liusie, and Gales]{manakul2023selfcheckgpt}
Potsawee Manakul, Adian Liusie, and Mark~JF Gales.
\newblock Selfcheckgpt: Zero-resource black-box hallucination detection for generative large language models.
\newblock \emph{arXiv preprint arXiv:2303.08896}, 2023.

\bibitem[Tian et~al.(2023)Tian, Mitchell, Zhou, Sharma, Rafailov, Yao, Finn, and Manning]{tian2023just}
Katherine Tian, Eric Mitchell, Allan Zhou, Archit Sharma, Rafael Rafailov, Huaxiu Yao, Chelsea Finn, and Christopher~D Manning.
\newblock Just ask for calibration: Strategies for eliciting calibrated confidence scores from language models fine-tuned with human feedback.
\newblock \emph{arXiv preprint arXiv:2305.14975}, 2023.

\bibitem[Xiong et~al.(2024)Xiong, Hu, Lu, Li, Fu, He, and Hooi]{xiong2024llmsexpressuncertaintyempirical}
Miao Xiong, Zhiyuan Hu, Xinyang Lu, Yifei Li, Jie Fu, Junxian He, and Bryan Hooi.
\newblock Can llms express their uncertainty? an empirical evaluation of confidence elicitation in llms, 2024.
\newblock URL \url{https://arxiv.org/abs/2306.13063}.

\bibitem[Yona et~al.(2024)Yona, Aharoni, and Geva]{yona2024can}
Gal Yona, Roee Aharoni, and Mor Geva.
\newblock Can large language models faithfully express their intrinsic uncertainty in words?
\newblock \emph{arXiv preprint arXiv:2405.16908}, 2024.

\bibitem[Belem et~al.(2024)Belem, Kelly, Steyvers, Singh, and Smyth]{belem2024perceptions}
Catarina~G Belem, Markelle Kelly, Mark Steyvers, Sameer Singh, and Padhraic Smyth.
\newblock Perceptions of linguistic uncertainty by language models and humans.
\newblock \emph{arXiv preprint arXiv:2407.15814}, 2024.

\bibitem[Zhu et~al.(2023)Zhu, Xu, Wang, Zhang, and Mao]{zhu2023calibration}
Chiwei Zhu, Benfeng Xu, Quan Wang, Yongdong Zhang, and Zhendong Mao.
\newblock On the calibration of large language models and alignment.
\newblock \emph{arXiv preprint arXiv:2311.13240}, 2023.

\bibitem[Guo et~al.(2025)Guo, Yang, Zhang, Song, Zhang, Xu, Zhu, Ma, Wang, Bi, et~al.]{guo2025deepseek}
Daya Guo, Dejian Yang, Haowei Zhang, Junxiao Song, Ruoyu Zhang, Runxin Xu, Qihao Zhu, Shirong Ma, Peiyi Wang, Xiao Bi, et~al.
\newblock Deepseek-r1: Incentivizing reasoning capability in llms via reinforcement learning.
\newblock \emph{arXiv preprint arXiv:2501.12948}, 2025.

\bibitem[Team(2024{\natexlab{a}})]{qwen3blog2024}
Qwen Team.
\newblock Qwen2 and qwen3: The new generation open models.
\newblock \url{https://qwenlm.github.io/blog/qwen3/}, 2024{\natexlab{a}}.
\newblock Accessed: 2025-05-12.

\bibitem[AI(2025)]{meta2025llama4}
Meta AI.
\newblock Llama 4: Multimodal intelligence.
\newblock \url{https://ai.meta.com/blog/llama-4-multimodal-intelligence/}, April 2025.
\newblock Accessed: 2025-05-10.

\bibitem[OpenAI(2024{\natexlab{a}})]{openai2024o3o4mini}
OpenAI.
\newblock Introducing o3 and o4 mini.
\newblock \url{https://openai.com/index/introducing-o3-and-o4-mini/}, 2024{\natexlab{a}}.
\newblock Accessed: 2025-05-12.

\bibitem[xAI(2024)]{xai2024grok3}
xAI.
\newblock Grok-3 and the next generation of reasoning models.
\newblock \url{https://x.ai/news/grok-3}, 2024.
\newblock Accessed: 2025-05-12.

\bibitem[Joshi et~al.(2017)Joshi, Choi, Weld, and Zettlemoyer]{joshi-etal-2017-triviaqa}
Mandar Joshi, Eunsol Choi, Daniel Weld, and Luke Zettlemoyer.
\newblock {T}rivia{QA}: A large scale distantly supervised challenge dataset for reading comprehension.
\newblock In Regina Barzilay and Min-Yen Kan, editors, \emph{Proceedings of the 55th Annual Meeting of the Association for Computational Linguistics (Volume 1: Long Papers)}, pages 1601--1611, Vancouver, Canada, July 2017. Association for Computational Linguistics.
\newblock \doi{10.18653/v1/P17-1147}.
\newblock URL \url{https://aclanthology.org/P17-1147/}.

\bibitem[Hendrycks et~al.(2020)Hendrycks, Burns, Basart, Zou, Mazeika, Song, and Steinhardt]{hendrycks2020measuring}
Dan Hendrycks, Collin Burns, Steven Basart, Andy Zou, Mantas Mazeika, Dawn Song, and Jacob Steinhardt.
\newblock Measuring massive multitask language understanding.
\newblock \emph{arXiv preprint arXiv:2009.03300}, 2020.

\bibitem[Wang et~al.(2024)Wang, Ma, Zhang, Ni, Chandra, Guo, Ren, Arulraj, He, Jiang, et~al.]{wang2024mmlu}
Yubo Wang, Xueguang Ma, Ge~Zhang, Yuansheng Ni, Abhranil Chandra, Shiguang Guo, Weiming Ren, Aaran Arulraj, Xuan He, Ziyan Jiang, et~al.
\newblock Mmlu-pro: A more robust and challenging multi-task language understanding benchmark.
\newblock In \emph{The Thirty-eight Conference on Neural Information Processing Systems Datasets and Benchmarks Track}, 2024.

\bibitem[Burns et~al.(2022)Burns, Ye, Klein, and Steinhardt]{burns2022discovering}
Collin Burns, Haotian Ye, Dan Klein, and Jacob Steinhardt.
\newblock Discovering latent knowledge in language models without supervision.
\newblock In \emph{International Conference on Learning Representations}, 2022.

\bibitem[Kadavath et~al.(2022)Kadavath, Conerly, Askell, Henighan, Drain, Perez, Schiefer, Hatfield-Dodds, DasSarma, Tran-Johnson, et~al.]{kadavath2022language}
Saurav Kadavath, Tom Conerly, Amanda Askell, Tom Henighan, Dawn Drain, Ethan Perez, Nicholas Schiefer, Zac Hatfield-Dodds, Nova DasSarma, Eli Tran-Johnson, et~al.
\newblock Language models (mostly) know what they know.
\newblock \emph{arXiv preprint arXiv:2207.05221}, 2022.

\bibitem[Farquhar et~al.(2024)Farquhar, Kossen, Kuhn, and Gal]{farquhar2024detecting}
Sebastian Farquhar, Jannik Kossen, Lorenz Kuhn, and Yarin Gal.
\newblock Detecting hallucinations in large language models using semantic entropy.
\newblock \emph{Nature}, 630\penalty0 (8017):\penalty0 625--630, 2024.

\bibitem[Guo et~al.(2017)Guo, Pleiss, Sun, and Weinberger]{guo2017calibration}
Chuan Guo, Geoff Pleiss, Yu~Sun, and Kilian~Q Weinberger.
\newblock On calibration of modern neural networks.
\newblock In \emph{International conference on machine learning}, pages 1321--1330. PMLR, 2017.

\bibitem[Mucs{\'a}nyi et~al.(2024)Mucs{\'a}nyi, Kirchhof, and Oh]{mucsanyi2024benchmarking}
B{\'a}lint Mucs{\'a}nyi, Michael Kirchhof, and Seong~Joon Oh.
\newblock Benchmarking uncertainty disentanglement: Specialized uncertainties for specialized tasks.
\newblock \emph{Advances in neural information processing systems}, 37:\penalty0 50972--51038, 2024.

\bibitem[Nixon et~al.(2019)Nixon, Dusenberry, Zhang, Jerfel, and Tran]{nixon2019measuring}
Jeremy Nixon, Michael~W Dusenberry, Linchuan Zhang, Ghassen Jerfel, and Dustin Tran.
\newblock Measuring calibration in deep learning.
\newblock In \emph{CVPR workshops}, volume~2, 2019.

\bibitem[Niculescu-Mizil and Caruana(2005)]{niculescu2005predicting}
Alexandru Niculescu-Mizil and Rich Caruana.
\newblock Predicting good probabilities with supervised learning.
\newblock In \emph{Proceedings of the 22nd international conference on Machine learning}, pages 625--632, 2005.

\bibitem[Grattafiori et~al.(2024)Grattafiori, Dubey, Jauhri, Pandey, Kadian, Al-Dahle, Letman, Mathur, Schelten, Vaughan, et~al.]{grattafiori2024llama}
Aaron Grattafiori, Abhimanyu Dubey, Abhinav Jauhri, Abhinav Pandey, Abhishek Kadian, Ahmad Al-Dahle, Aiesha Letman, Akhil Mathur, Alan Schelten, Alex Vaughan, et~al.
\newblock The llama 3 herd of models.
\newblock \emph{arXiv preprint arXiv:2407.21783}, 2024.

\bibitem[Liu et~al.(2024{\natexlab{a}})Liu, Feng, Xue, Wang, Wu, Lu, Zhao, Deng, Zhang, Ruan, et~al.]{liu2024deepseek}
Aixin Liu, Bei Feng, Bing Xue, Bingxuan Wang, Bochao Wu, Chengda Lu, Chenggang Zhao, Chengqi Deng, Chenyu Zhang, Chong Ruan, et~al.
\newblock Deepseek-v3 technical report.
\newblock \emph{arXiv preprint arXiv:2412.19437}, 2024{\natexlab{a}}.

\bibitem[Yang et~al.(2024)Yang, Yang, Zhang, Hui, Zheng, Yu, Li, Liu, Huang, Wei, et~al.]{yang2024qwen2}
An~Yang, Baosong Yang, Beichen Zhang, Binyuan Hui, Bo~Zheng, Bowen Yu, Chengyuan Li, Dayiheng Liu, Fei Huang, Haoran Wei, et~al.
\newblock Qwen2. 5 technical report.
\newblock \emph{arXiv preprint arXiv:2412.15115}, 2024.

\bibitem[Jiang(2024)]{jiang2024identifying}
Fengqing Jiang.
\newblock Identifying and mitigating vulnerabilities in llm-integrated applications.
\newblock Master's thesis, University of Washington, 2024.

\bibitem[Team et~al.(2024{\natexlab{a}})Team, Mesnard, Hardin, Dadashi, Bhupatiraju, Pathak, Sifre, Rivi{\`e}re, Kale, Love, et~al.]{team2024gemma}
Gemma Team, Thomas Mesnard, Cassidy Hardin, Robert Dadashi, Surya Bhupatiraju, Shreya Pathak, Laurent Sifre, Morgane Rivi{\`e}re, Mihir~Sanjay Kale, Juliette Love, et~al.
\newblock Gemma: Open models based on gemini research and technology.
\newblock \emph{arXiv preprint arXiv:2403.08295}, 2024{\natexlab{a}}.

\bibitem[OpenAI(2024{\natexlab{b}})]{openai2024gpt4o}
OpenAI.
\newblock Introducing gpt-4o: Openai’s new omni model.
\newblock \url{https://openai.com/index/hello-gpt-4o/}, 2024{\natexlab{b}}.
\newblock Accessed: 2025-05-12.

\bibitem[OpenAI(2024{\natexlab{c}})]{openai2024gpt41}
OpenAI.
\newblock Gpt-4.1 overview.
\newblock \url{https://openai.com/index/gpt-4-1/}, 2024{\natexlab{c}}.
\newblock Accessed: 2025-05-12.

\bibitem[Anthropic(2024)]{anthropic2024claude3}
Anthropic.
\newblock Introducing the claude 3 model family.
\newblock \url{https://www.anthropic.com/news/claude-3-family}, 2024.
\newblock Accessed: 2025-05-12.

\bibitem[Team et~al.(2024{\natexlab{b}})Team, Georgiev, Lei, Burnell, Bai, Gulati, Tanzer, Vincent, Pan, Wang, et~al.]{team2024gemini}
Gemini Team, Petko Georgiev, Ving~Ian Lei, Ryan Burnell, Libin Bai, Anmol Gulati, Garrett Tanzer, Damien Vincent, Zhufeng Pan, Shibo Wang, et~al.
\newblock Gemini 1.5: Unlocking multimodal understanding across millions of tokens of context.
\newblock \emph{arXiv preprint arXiv:2403.05530}, 2024{\natexlab{b}}.

\bibitem[Team(2024{\natexlab{b}})]{qwen3report2024}
Qwen Team.
\newblock Qwen3 technical report.
\newblock \texttt{GitHub Repository}, 2024{\natexlab{b}}.
\newblock URL \url{https://github.com/QwenLM/Qwen3/blob/main/Qwen3_Technical_Report.pdf}.
\newblock \url{https://github.com/QwenLM/Qwen3}.

\bibitem[Abdi(2007)]{abdi2007kendall}
Herv{\'e} Abdi.
\newblock The kendall rank correlation coefficient.
\newblock \emph{Encyclopedia of measurement and statistics}, 2:\penalty0 508--510, 2007.

\bibitem[Kim et~al.(2024)Kim, Liao, Vorvoreanu, Ballard, and Vaughan]{kim2024m}
Sunnie~SY Kim, Q~Vera Liao, Mihaela Vorvoreanu, Stephanie Ballard, and Jennifer~Wortman Vaughan.
\newblock " i'm not sure, but...": Examining the impact of large language models' uncertainty expression on user reliance and trust.
\newblock In \emph{Proceedings of the 2024 ACM Conference on Fairness, Accountability, and Transparency}, pages 822--835, 2024.

\bibitem[Lin et~al.(2023)Lin, Trivedi, and Sun]{lin2023generating}
Zhen Lin, Shubhendu Trivedi, and Jimeng Sun.
\newblock Generating with confidence: Uncertainty quantification for black-box large language models.
\newblock \emph{arXiv preprint arXiv:2305.19187}, 2023.

\bibitem[Azaria and Mitchell(2023)]{azaria2023internal}
Amos Azaria and Tom Mitchell.
\newblock The internal state of an llm knows when it's lying.
\newblock In \emph{Findings of the Association for Computational Linguistics: EMNLP 2023}, pages 967--976, 2023.

\bibitem[Li et~al.(2023)Li, Patel, Vi{\'e}gas, Pfister, and Wattenberg]{li2023inference}
Kenneth Li, Oam Patel, Fernanda Vi{\'e}gas, Hanspeter Pfister, and Martin Wattenberg.
\newblock Inference-time intervention: Eliciting truthful answers from a language model.
\newblock In \emph{Advances in Neural Information Processing Systems}, 2023.

\bibitem[Liu et~al.(2024{\natexlab{b}})Liu, Pan, Li, and Chen]{liu2024uncertainty}
Linyu Liu, Yu~Pan, Xiaocheng Li, and Guanting Chen.
\newblock Uncertainty estimation and quantification for llms: A simple supervised approach.
\newblock \emph{arXiv preprint arXiv:2404.15993}, 2024{\natexlab{b}}.

\bibitem[Geng et~al.(2023)Geng, Cai, Wang, Koeppl, Nakov, and Gurevych]{geng2023survey}
Jiahui Geng, Fengyu Cai, Yuxia Wang, Heinz Koeppl, Preslav Nakov, and Iryna Gurevych.
\newblock A survey of confidence estimation and calibration in large language models.
\newblock \emph{arXiv preprint arXiv:2311.08298}, 2023.

\bibitem[Huang et~al.(2024)Huang, Yang, Zhang, Lee, and Wu]{huang2024survey}
Hsiu-Yuan Huang, Yutong Yang, Zhaoxi Zhang, Sanwoo Lee, and Yunfang Wu.
\newblock A survey of uncertainty estimation in llms: Theory meets practice.
\newblock \emph{arXiv preprint arXiv:2410.15326}, 2024.

\bibitem[Liu et~al.(2025)Liu, Chen, Da, Chen, Lin, and Wei]{liu2025uncertainty}
Xiaoou Liu, Tiejin Chen, Longchao Da, Chacha Chen, Zhen Lin, and Hua Wei.
\newblock Uncertainty quantification and confidence calibration in large language models: A survey.
\newblock \emph{arXiv preprint arXiv:2503.15850}, 2025.

\end{thebibliography}
